\documentclass{article}
\usepackage{iclr2027_conference,times}
\usepackage[utf8]{inputenc}
\usepackage[T1]{fontenc}
\usepackage{amsmath,amssymb}
\usepackage{booktabs,multirow,makecell,array,tabularx}
\usepackage{graphicx}
\usepackage{xcolor}
\usepackage{colortbl}
\usepackage{caption}
\usepackage{float}
\usepackage{listings}
\usepackage[most]{tcolorbox}
\tcbuselibrary{skins,breakable,listings}
\usepackage{fancyvrb}
\usepackage{algorithm}
\usepackage{algpseudocode}
\usepackage{hyperref}
\usepackage{url}

\newcommand{\GraphSkillAA}{\textsc{GraphSkillAA}}
\newcommand{\SkillOpt}{\textsc{SkillOpt}}

\newcommand{\hrange}[2]{#1$_{\pm #2}$}
\definecolor{gainred}{HTML}{C62828}
\definecolor{dropgreen}{HTML}{2E7D32}
\definecolor{tableheadergray}{HTML}{EBEBEB}
\newcommand{\gain}[1]{\textsuperscript{\textcolor{gainred}{+#1}}}
\newcommand{\drop}[1]{\textsuperscript{\textcolor{dropgreen}{-#1}}}
\newcommand{\pp}{\,pp}
\newcommand{\lightrowrule}{\arrayrulecolor{black!18}\hline\arrayrulecolor{black}}

\newtcolorbox{badcasebox}[1]{
  breakable,
  colback=blue!4!white,
  colframe=blue!45!black,
  colbacktitle=blue!4!white,
  coltitle=black,
  fonttitle=\bfseries,
  fontupper=\small,
  title={#1},
  titlerule=0pt,
  arc=1.5mm,
  boxrule=0.6pt,
  left=2mm,
  right=2mm,
  top=1.2mm,
  bottom=1.2mm,
  before skip=5pt,
  after skip=5pt
}

\newtcblisting{promptbox}[1][]{
  enhanced,
  listing only,
  colframe=black!72,
  colback=black!3,
  colbacktitle=black!3,
  coltitle=black,
  fonttitle=\small\bfseries\rmfamily,
  fontupper=\small\rmfamily,
  title={#1},
  titlerule=0pt,
  arc=1.5mm,
  outer arc=1.5mm,
  boxrule=0.7pt,
  top=1.2mm,
  bottom=1.2mm,
  left=2mm,
  right=2mm,
  before skip=1em,
  after skip=1em,
  listing options={
    basicstyle=\small\rmfamily,
    breaklines=true,
    breakatwhitespace=true,
    columns=flexible,
    keepspaces=true,
    showstringspaces=false
  }
}

\title{\scalebox{0.93}{\mbox{GraphSkillAA: Attribution-Guided Skill-Graph}}\protect\\
\scalebox{0.93}{\mbox{Updating with Targeted Validation and Rollback}}}
\author{
Ziqiao Shang\textsuperscript{\rm 1,2},
Lingyue Ge\textsuperscript{\rm 1,2},
Lan-Zhe Guo\textsuperscript{\rm 1,2}\textsuperscript{*}\\
\textsuperscript{\rm 1}National Key Laboratory for Novel Software Technology, Nanjing University, Nanjing, China\\
\textsuperscript{\rm 2}School of Intelligence Science and Technology, Nanjing University, Suzhou, China\\
\small{\textsuperscript{*}Corresponding author.}\\
\small{\textbf{Emails:} shangzq@lamda.nju.edu.cn\hspace{1em}gely@lamda.nju.edu.cn\hspace{1em}guolz@nju.edu.cn}\\
\small{\textbf{Code:} \url{https://github.com/Ziqiao-Shang/SkillAA}}
}
\iclrfinalcopy

\begin{document}
\raggedbottom
\maketitle

\begin{abstract}
External skills provide domain knowledge and procedures without updating model parameters, but flat text makes their applicability, dependencies, and composition difficult to express. Existing skill graphs organize rules into nodes and relations mainly for retrieval and composition. They give less attention to what should change after a failure, which examples should be retested, and which edits should be reverted. We introduce \GraphSkillAA{} (\emph{Graph-Skill Abductive Attribution}), which connects rule use, error localization, and update validation through the same skill graph. Nodes record applicability, procedures, and exclusion conditions, while edges describe prerequisite and enhancement relations. The student records used nodes and edges. The teacher contrasts related successes and failures to identify the fields or relations to repair, keeping changes as small as possible. Execution lapses or insufficient evidence leave the graph unchanged. A Local Gate retests affected examples, and a Big Gate checks the merged graph on the complete update pool. Rejected edits are rolled back, and model parameters remain frozen throughout. With GPT-5.6-sol, \GraphSkillAA{} reaches 81.5\%, 66.7\%, and 91.2\% on SearchQA, LiveMath, and DocVQA, respectively, and attains the highest mean in every main setting. These results show that graph structure supports skill selection and composition while also providing explicit targets for repair, retesting, and rollback.
\end{abstract}

\begin{figure}[ht]
\vspace{-3pt}
\centering
\begin{minipage}{\dimexpr\textwidth-6pc\relax}
\centering
\includegraphics[width=\linewidth]{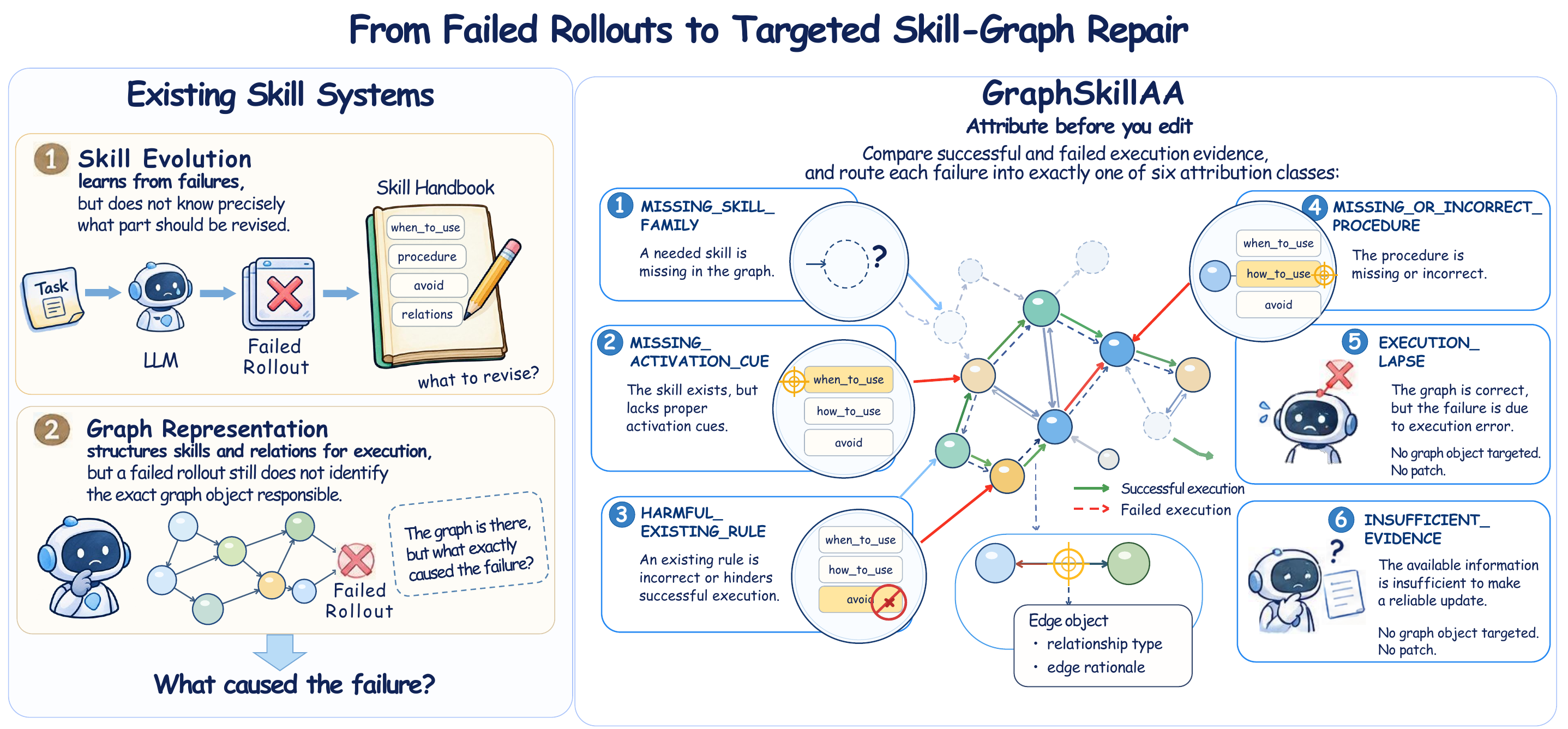}
\setlength{\abovecaptionskip}{3pt}
\setlength{\belowcaptionskip}{0pt}
\caption{\GraphSkillAA{}: first localize the cause of failure, then update the skill locally. \emph{Left:} Existing methods can update skills or organize a skill graph, but have difficulty determining exactly what should change after a failure. \emph{Right:} \GraphSkillAA{} contrasts successful and failed execution traces, decides whether an update is warranted, localizes a specific skill field or relation, and makes only an evidence-supported local edit. If the failure is an execution lapse or the evidence is insufficient, the skill graph remains unchanged.}
\label{fig:intro-attribution}
\end{minipage}
\vspace{-4pt}
\end{figure}

\section{Introduction}

Frozen language models increasingly use external tools, documents, workflows, and skills to acquire domain procedures without changing model parameters~\citep{schick2023toolformer,patil2023gorilla,qin2024toolllm}. External skills are easy to read, transfer, and revise, but adding instructions does not necessarily improve performance. An incorrect answer may reflect missing knowledge, failure to select a suitable rule, faulty rule content or relations, or an execution lapse. Rewriting skills from the final error alone may fix that example while harming previously correct answers. Skill optimization therefore needs to identify which rules or relations should change, then compare before--after results to determine whether the gains outweigh newly introduced errors.

A graph representation provides a natural interface for this problem. Node fields can separate when a skill applies, how it executes, and when it should be avoided; stable IDs make these fields addressable; typed edges expose prerequisite and enhancement relations; and dependency paths describe how several skills compose. Recent systems use these properties to retrieve relevant skills and construct executable bundles~\citep{skillrouter2026,graphofskills2026,skilldag2026,hyperskill2026,caskg2026,skillgraph2026}. However, existing methods typically use the graph only before inference to retrieve and compose skills. Once the model answers incorrectly, the update stage often treats the skill library as an undifferentiated text artifact: it has difficulty determining whether the problem lies in a skill's applicability conditions, execution steps, or relations to other skills, and it does not use graph dependencies to decide which previously correct cases should be retested. Skill graphs therefore also need to support edit localization after failures and the assessment of both repair gains and newly introduced errors.

We introduce \GraphSkillAA{} (\emph{Graph-Skill Abductive Attribution}). Skills are stored as nodes and relations with stable IDs, allowing the teacher to specify the fields or edges to edit. After a failure, the teacher compares related successful cases to determine whether the problem concerns a missing rule, applicability conditions, procedure steps, or node relations, and proposes a repair with as few changes as possible. Insufficient evidence for a skill edit produces \texttt{NO\_PATCH}. Related edits form complete repair groups, which are retested on affected examples before the accepted groups are merged for whole-graph validation. Rejected edits are reverted while other effective edits can be retained.

Across SearchQA, LiveMath, and DocVQA, \GraphSkillAA{} attains the highest mean in every main setting. With GPT-5.6-sol, accuracies are 81.5\%, 66.7\%, and 91.2\%, respectively. With all node content retained, removing edges lowers GPT-5.6-sol accuracy by $4.0$, $4.6$, and $1.2\pp$, demonstrating the contribution of relations beyond node representation. Shuffling initial edit targets further shows that diagnosed targets exceed random type-compatible targets by $3.2$, $6.0$, and $1.5\pp$ on test accuracy. Together with retrieval coverage and two-level validation analyses, these results connect graph structure to skill composition, repair localization, and update checking.

Our contributions are:
\begin{enumerate}
\item \textbf{A skill graph for localized updating:} stable IDs identify rule fields and compositional relations, giving each edit an explicit target.
\item \textbf{Object-level abductive attribution:} repairs are proposed from success--failure contrasts, while execution lapses or insufficient evidence preserve the graph.
\item \textbf{Two-level validation and rollback:} interdependent edits are retested as groups before the merged graph is validated, allowing edits to be retained or reverted together.
\item \textbf{Evaluation across models and tasks:} experiments on open-domain QA, mathematical reasoning, and document understanding compare the complete method and its components.
\end{enumerate}

\section{Related Work}

\subsection{Graph-Structured Skills}
Graphs organize document evidence, memory, and tool relations: ControlLLM models tool dependencies; CRAFT builds and retrieves specialized toolsets; ToolRerank uses hierarchical relations for reranking; and GraphRAG and HippoRAG connect distributed knowledge through relational graphs~\citep{liu2023controlllm,yuan2024craft,zheng2024toolrerank,edge2024graphrag,gutierrez2024hipporag}. Skill libraries extend this idea: SkillRouter retrieves from skill-body signals; Graph-of-Skills returns dependency-aware bundles; SkillGraph and SkillDAG evolve typed relations; and HyperSkill and CaSKG model higher-order or directed composition~\citep{skillrouter2026,graphofskills2026,skillgraph2026,skilldag2026,hyperskill2026,caskg2026}. These methods show that relations aid compositional execution and mainly focus on graph construction and retrieval. \GraphSkillAA{} further uses the same graph objects for attribution, targeted updating, validation, and rollback.

\subsection{Skill Evolution and Controlled Optimization}
Optimization of prompts and skill documents has progressed from searching candidate text to revision based on execution trajectories. ProTeGi, OPRO, TextGrad, and GEPA optimize prompts or agent configurations from critiques, textual feedback, or trajectories~\citep{pryzant2023protegi,yang2024opro,yuksekgonul2024textgrad,agrawal2025gepa}. Skill systems similarly move from experience accumulation in Voyager and ExpeL to reusable workflows, skills, or memories in Agent Workflow Memory, SkillWeaver, ReasoningBank, and Trace2Skill~\citep{wang2023voyager,zhao2024expel,wang2024awm,zheng2025skillweaver,reasoningbank2025,trace2skill2026}. SkillsBench and related surveys define skills as reusable procedures with applicability conditions~\citep{skillsbench2026,agentskillssurvey2026}, while SkillOpt improves skill documents through bounded edits and multilevel validation~\citep{skillopt2026}. These approaches mainly optimize text or complete candidates, making it difficult to edit, validate, and revert individual fields or skill relations separately.

\section{Method}
\label{sec:method}

\subsection{Overview: Skill Optimization as Graph-State Editing}
\label{sec:method-overview}

\begin{figure*}[t]
\centering
\includegraphics[width=0.99\textwidth]{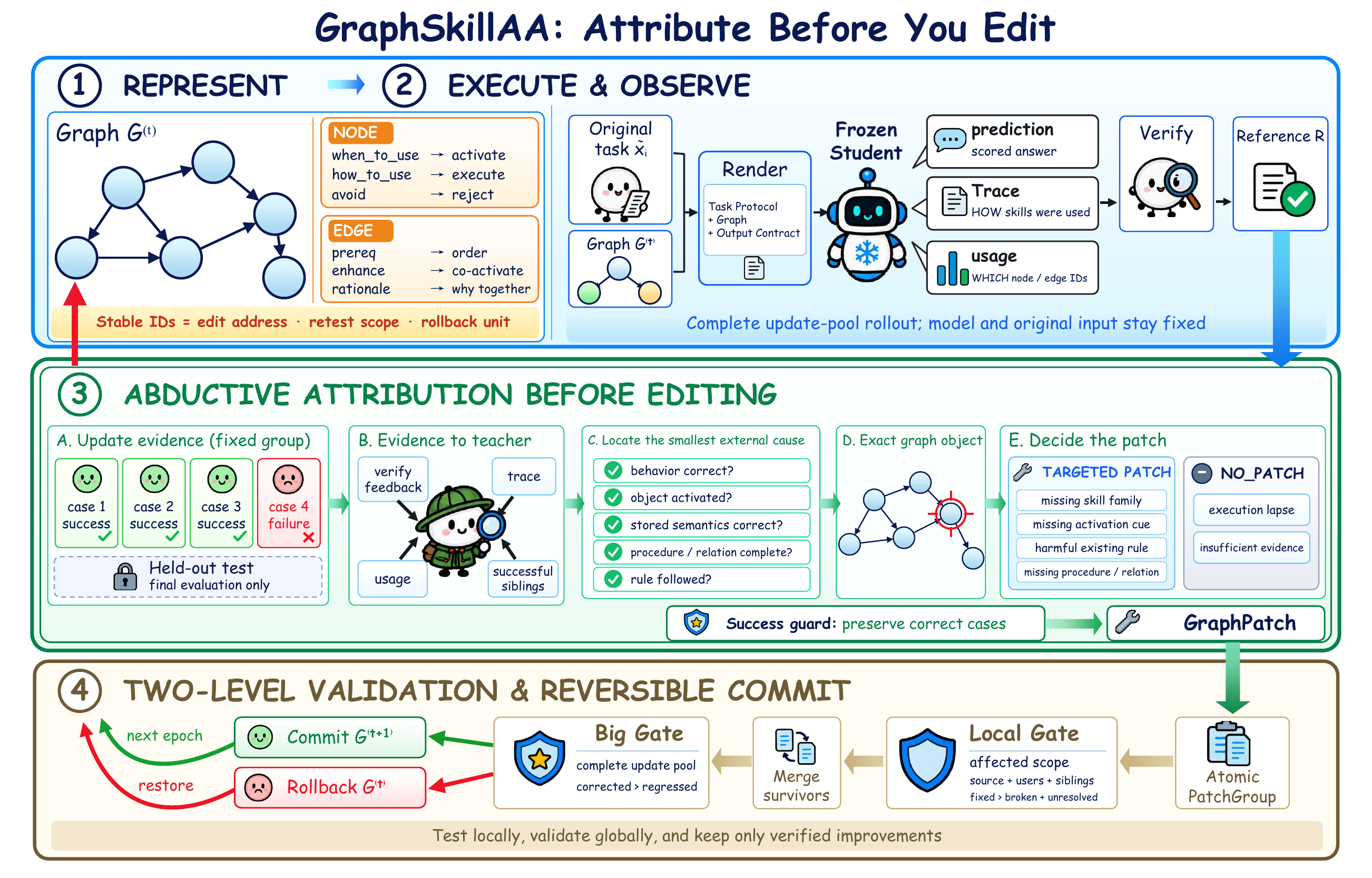}
\setlength{\abovecaptionskip}{3pt}
\setlength{\belowcaptionskip}{0pt}
\caption{The \GraphSkillAA{} execution and update pipeline. The student completes tasks using the skill graph. The teacher contrasts successful and failed records to identify edit targets and organize related edits into repair groups. The Local Gate checks each group on affected examples, and the Big Gate checks the merged graph on the complete update set. After whole-graph acceptance, the graph and per-example results enter the next epoch together.}
\label{fig:framework-overview}
\end{figure*}

\GraphSkillAA{} builds on the outer optimization framework of \SkillOpt{}~\citep{skillopt2026}, retaining its model access, data and rollout organization, and epoch loop while converting the flat skill document into an addressable skill graph. This representation allows updates to target specific rules or relations instead of rewriting the entire document. Figure~\ref{fig:framework-overview} shows the overall \GraphSkillAA{} workflow. The system takes an initial skill specification, an update pool, frozen student and teacher models, and an optimization budget $E$ as input, and returns the final committed graph $G^{(E)}$. During optimization, only the skill graph changes; model parameters, original task inputs, and the output contract remain fixed, and $\mathcal{D}_{\mathrm{test}}$ never enters updating. Appendix~\ref{app:skillaa-overview} gives the complete algorithm.

Let the frozen student $M_{\mathrm{student}}$ execute tasks and the frozen teacher $M_{\mathrm{teacher}}$ attribute failures and propose edits. The update pool is
\[
\mathcal{D}_{\mathrm{update}}=\{(x_i,y_i)\}_{i=1}^{N},
\]
where $i$ indexes one of $N$ examples and $y_i$ is a gold answer or verifiable feedback. The final-evaluation set $\mathcal{D}_{\mathrm{test}}$ is disjoint. The system first initializes a task skill document. In our experiments, the best document optimized by \SkillOpt{} supplies this initialization. An LLM then semantically converts the document into reusable rule nodes and edges expressing prerequisites, enhancement, and joint use, yielding $G^{(0)}$. Runs share this initial graph and subsequently update it using execution feedback. Here, $G^{(t)}$ denotes the graph committed at epoch $t$. Appendix~\ref{app:skill-to-graph} gives the conversion and an example.

A deterministic renderer combines the current graph with the fixed task protocol and output contract:
\[
P^{(t)}
=\operatorname{Render}\!\left(
\mathrm{TaskProtocol},\,G^{(t)},\,\mathrm{OutputContract}
\right).
\]
$\mathrm{TaskProtocol}$ fixes task requirements and $\mathrm{OutputContract}$ the answer format; only $G^{(t)}$ changes. \textsc{Render} serializes the active nodes and edges in a fixed order.

Table~\ref{tab:update-stages} summarizes the stages and their inputs and outputs.

\begin{table}[H]
\centering
\caption{The four stages of each \GraphSkillAA{} update epoch and their inputs and outputs.}
\label{tab:update-stages}
\footnotesize
\renewcommand{\arraystretch}{1.14}
\setlength{\tabcolsep}{4pt}
\begin{tabularx}{\textwidth}{@{}>{\raggedright\arraybackslash}p{0.16\textwidth}>{\raggedright\arraybackslash}X>{\raggedright\arraybackslash}X@{}}
\toprule
\rowcolor{tableheadergray}
\textbf{Stage} & \textbf{Input} & \textbf{Output and next step} \\
\midrule
Localize the problem & Current graph, failed executions, related successful executions, and scoring feedback & Initial proposals listing all edit targets and source evidence for joint synthesis, or \texttt{NO\_PATCH} \\
\lightrowrule
Form repair groups & Initial proposals and related success--failure evidence & Structurally valid repair groups that can be retained or reverted as units, passed to the Local Gate \\
\lightrowrule
Local validation & Epoch-start graph, each repair group, affected examples, and pre-edit results & Accepted repair groups, passed to merging \\
\lightrowrule
Whole-graph validation & Merged candidate graph, complete update set, and pre-edit results & Commit the new graph and its results, or retain the epoch-start graph and results \\
\bottomrule
\end{tabularx}
\end{table}

\subsection{Addressable Skill Graph}
\label{sec:skill-graph}

\GraphSkillAA{} treats the graph as persistent state for selection, editing, validation, and restoration. The skill library is $G=(\mathrm{Nodes},\mathrm{Edges})$. A node
\[
v=\bigl(
\mathrm{id},\,
\mathrm{title},\,
\texttt{when\_to\_use},\,
\texttt{how\_to\_use},\,
\texttt{avoid}
\bigr)
\]
denotes a conditional reusable procedure: \texttt{when\_to\_use} states applicability, \texttt{how\_to\_use} specifies execution, and \texttt{avoid} records inapplicable situations or actions to avoid. Conceptually,
\[
\operatorname{Use}(v,x)\approx
\operatorname{Match}\!\left(x,\texttt{when\_to\_use}(v)\right)
\land
\neg\operatorname{Conflict}\!\left(x,\texttt{avoid}(v)\right).
\]
Here, $\approx$ denotes a conceptual judgment, and \textsc{Match} and \textsc{Conflict} are the frozen student's in-context applicability and exclusion judgments. Separating activation, execution, and exclusion makes these failure surfaces independently addressable.

An edge
\[
e=\bigl(
\mathrm{source},\,
\mathrm{target},\,
\mathrm{type},\,
\mathrm{strength},\,
\mathrm{rationale}
\bigr)
\]
describes composition. Source and target specify direction; \textsc{prereq} is required, \textsc{enhance} is beneficial, and \texttt{co\_occur} records joint use without imposing an execution order. Strength and rationale record priority and evidence.

\paragraph{Role of graph structure.}
Nodes describe individual skills; edges describe how skills work together. With a node list alone, the model must infer ordering and supplementary checks from the rules. A graph stores these links separately: \texttt{prereq} identifies what must precede a skill, \texttt{enhance} identifies how one skill assists another, and \texttt{co\_occur} records skills used together. The \texttt{rationale} explains why a relation holds and when it applies. Edges sharing endpoints connect these local relations into skill compositions without repeating every relation inside each node. The skill graph and example executions are illustrated in Appendix~\ref{app:searchqa-graph-workflow}.

\begin{badcasebox}{SearchQA example: answer type and granularity}
\textbf{Question:} ``Using a nuclear generator, the Nautilus was what type of vessel?''

\smallskip
\textbf{Q011---Identify the answer type:} Recognize that the question asks for a vessel type.\\
\textbf{Q029---Check answer granularity:} Return the specific category supported by the evidence. Here, the answer should be \emph{submarine}, rather than merely \emph{ship}.

\smallskip
\textbf{Relation:} $\mathrm{Q011}\xrightarrow{\texttt{enhance}}\mathrm{Q029}$.\\
The answer type identified by Q011 helps Q029 judge whether the answer is specific enough. The nodes describe the individual checks; the edge specifies how they work together.
\end{badcasebox}

The graph also makes relations editable objects. When two rules are individually correct but their connection has the wrong direction or explanation, the relation can be revised while preserving both rules. Logged node and edge IDs then identify affected examples, while node--edge dependencies identify operations that must be retained or restored together. Graph structure thus supports composition during execution and relation repair and local validation after failure.

At inference, \GraphSkillAA{} renders all active nodes and edges in stable order, without external embedding-based Top-$K$ pruning (Appendix~\ref{app:retrieval}). The student selects nodes from \texttt{when\_to\_use} and \texttt{avoid}, executes \texttt{how\_to\_use}, and follows edge semantics. Stable IDs identify the same objects during selection, attribution-guided editing, and impact analysis. Interdependent edits form repair groups that can be retained or reverted together.

\subsection{Execution Evidence and Abductive Attribution}
\label{sec:localization}

A failed answer reveals that the current execution is inadequate, but not which persistent state, if any, should change. The cause may be a missing skill, missed activation, harmful rule, incomplete procedure or relation, or a one-off execution lapse; editing before diagnosis can therefore modify the wrong object. \GraphSkillAA{} instead contrasts the failure with related successes, seeks a reusable graph-level explanation supported by the evidence, and routes the initial repair to its corresponding edit surface.

To support this diagnosis, attribution observes both outcome correctness and graph participation. On each original input, the student returns
\[
\begin{gathered}
(\mathrm{prediction}_i^{(t)},\,
 \mathrm{trace}_i^{(t)},\,
 \mathrm{usage}_i^{(t)})
=M_{\mathrm{student}}(x_i;P^{(t)}),\\
\mathrm{correct}_i^{(t)}
=\operatorname{Verify}(\mathrm{prediction}_i^{(t)},y_i).
\end{gathered}
\]
\texttt{prediction} is the submitted answer. \texttt{trace} records the semantic path, for example, \emph{Q3 determines relation direction, E2 invokes Q7, and Q7 normalizes the answer}; \texttt{usage} lists the same path by stable ID, for example, \texttt{used\_nodes=[Q3,Q7]} and \texttt{used\_edges=[E2]}. The former explains \emph{how} skills were used for attribution; the latter identifies \emph{which} objects may be affected. The program checks their IDs and agreement, marking missing or invalid usage records as unknown (Appendix~\ref{app:update-contract}).

The verifier compares only \texttt{prediction} with $y_i$, yielding $\mathrm{correct}_i^{(t)}\in\{0,1\}$. A full-pool rollout---predictions, traces, usage, and correctness---forms the \emph{reference} $R^{(t)}$ for same-example candidate comparisons.

Before optimization, \GraphSkillAA{} partitions only the update pool into fixed, disjoint quadruples using task-type constraints and question-text similarity:
\[
\mathrm{Group}_m=
\{\mathrm{update}_{m,1},\mathrm{update}_{m,2},
  \mathrm{update}_{m,3},\mathrm{update}_{m,4}\}.
\]
Each complete group contains four update examples; any remaining examples are retained in the final batch. Successful members provide local contrast paths, and members of relevant groups enter the Local Gate.

For each failure, the teacher combines verifier feedback, \texttt{trace}, validated \texttt{usage}, and successful group members. It locates the earliest observable error, then follows this diagnostic chain:
\[
\begin{gathered}
\text{behavior covered?}
\rightarrow
\text{relevant object activated?}
\rightarrow
\text{stored semantics correct?}\\
\rightarrow
\text{steps or relations complete?}
\rightarrow
\text{rule followed?}
\end{gathered}
\]
Table~\ref{tab:attribution-routing} maps the diagnosis to six routes. The first four constrain the initial object-level \textsc{GraphPatch}; execution lapse and insufficient evidence return \texttt{NO\_PATCH}. Here, \emph{abductive} attribution infers the graph-level defect from contrasting execution evidence and proposes a repair with as few changes as possible. Initial proposals list all target nodes, fields, and relations, including dependent operations. Appendix~\ref{app:edit-routing} gives the diagnostic order, and Appendix~\ref{app:prompts} collects prompts and examples.

\begin{table*}[t]
\centering
\footnotesize
\renewcommand{\arraystretch}{1.14}
\setlength{\tabcolsep}{4.0pt}
\caption{Attribution-to-update routing. Each route constrains its initial graph edit.}
\label{tab:attribution-routing}
\begin{tabular}{
@{}
>{\raggedright\arraybackslash}p{0.20\textwidth}
>{\raggedright\arraybackslash}p{0.31\textwidth}
>{\raggedright\arraybackslash}p{0.41\textwidth}
@{}
}
\toprule
\rowcolor{tableheadergray}
\multicolumn{1}{c}{\textbf{Root-cause code}}
&
\multicolumn{1}{c}{\textbf{Failure condition}}
&
\multicolumn{1}{c}{\textbf{Authorized update}}
\\
\midrule
\makecell[tl]{\scriptsize\texttt{MISSING\_}\\\scriptsize\texttt{SKILL\_FAMILY}}
&
No node covers the required independent, reusable procedure.
&
Atomically add a specialist node and its connecting \texttt{enhance} edge.
\\
\lightrowrule
\makecell[tl]{\scriptsize\texttt{MISSING\_}\\\scriptsize\texttt{ACTIVATION\_CUE}}
&
A correct, applicable skill exists but is absent from validated \texttt{usage}.
&
Extend only its \texttt{when\_to\_use}.
\\
\lightrowrule
\makecell[tl]{\scriptsize\texttt{HARMFUL\_}\\\scriptsize\texttt{EXISTING\_RULE}}
&
A used node or edge contains harmful semantics.
&
Replace only the harmful field; atomically replace a faulty edge.
\\
\lightrowrule
\makecell[tl]{\scriptsize\texttt{MISSING\_OR\_}\\\scriptsize\texttt{INCORRECT\_}\\\scriptsize\texttt{PROCEDURE}}
&
The correct node is active but lacks a step, exclusion boundary, or relation.
&
Complete only \texttt{how\_to\_use} or \texttt{avoid}, or add the required \texttt{prereq}/\texttt{enhance} edge.
\\
\lightrowrule
\makecell[tl]{\scriptsize\texttt{EXECUTION\_LAPSE}}
&
The graph is correct and active, but execution diverges.
&
Return \texttt{NO\_PATCH}; do not persist a one-off lapse.
\\
\lightrowrule
\makecell[tl]{\scriptsize\texttt{INSUFFICIENT\_}\\\scriptsize\texttt{EVIDENCE}}
&
Evidence cannot localize a field or edge.
&
Return \texttt{NO\_PATCH}; preserve the graph.
\\
\bottomrule
\end{tabular}
\end{table*}

\subsection{Targeted Updates and Two-Level Rollback}
\label{sec:repair-groups}

Attribution specifies the edit targets, and joint synthesis coordinates connected proposals within that scope using the same failure evidence and related successes. For example, a new node and its connecting edge are proposed together and form one atomic \textsc{PatchGroup} $J_k$. Each group records its source evidence and is retained or reverted as a unit. Structural checks reject invalid IDs, endpoints, conflicting relations, or cycles in \texttt{prereq} before local validation. Teacher updates modify node fields and \texttt{prereq}/\texttt{enhance} relations, while \texttt{co\_occur} retains its initial settings.

For example, adding Q029 entails proposing both the granularity check and its connecting edge E007. The Local Gate evaluates the new rule and its connection as one repair group. If the group fails, both node and edge are removed, while other effective repairs can survive. Rollback thus restores a complete change to skill composition rather than an isolated text fragment.

A targeted patch may still break other users of the same object, and individually useful patches may conflict after merging. The Local Gate therefore tests each atomic group on its affected scope; the Big Gate tests all survivors as one executable graph before epoch-level commitment. This ties validation cost and rollback granularity to graph structure.

\paragraph{Local Gate: graph-scoped retesting.}
\label{sec:local-validation}
Each group $J_k$ is applied to a copy of the epoch-start graph $G^{(t)}$ and compared with the same pre-edit reference $R^{(t)}$. \GraphSkillAA{} reruns the update examples selected by its affected scope:
\[
\mathrm{Retest}_k=
\operatorname{UpdateMembers}\!\left(
\mathcal Q_{\mathrm{src}}^k
\cup\mathcal Q_{\mathrm{use}}^k
\cup\mathcal Q_{\mathrm{unk}}
\right).
\]
$\mathcal Q_{\mathrm{src}}^k$ contains source groups; $\mathcal Q_{\mathrm{use}}^k$ contains groups whose validated \texttt{usage} includes the edited node or, for an existing edge, the edge or either endpoint; and $\mathcal Q_{\mathrm{unk}}$ contains groups with invalid or missing \texttt{usage}. $\operatorname{UpdateMembers}$ expands these groups to examples. For a new edge between existing nodes, the users of either endpoint enter $\mathcal Q_{\mathrm{use}}^k$. For a group that only adds a specialist node and its connecting edge, the retest scope starts from source and unknown-usage groups. Appendix~\ref{app:local-gate-details} gives the full scope rules.

Let $n_{\mathrm{fixed}}^k$, $n_{\mathrm{broken}}^k$, and $n_{\mathrm{unresolved}}^k$ count corrected errors, broken successes, and unresolved source failures. Retain the group only if
\[
n_{\mathrm{fixed}}^k
>
n_{\mathrm{broken}}^k+n_{\mathrm{unresolved}}^k.
\]
A rejected group that meets the refinement conditions may be revised once within its original target scope and retested against the same reference. If it fails again or is not eligible for refinement, all of $J_k$ is rolled back, restoring old fields and jointly reversing new-node--edge pairs or edge replacements. Graph objects thus determine retest scope, and structural dependencies determine rollback units (Appendix~\ref{app:local-gate-details}).

\paragraph{Big Gate: epoch-level commit.}
\label{sec:whole-graph-validation}
Local survivors are merged in proposal order into the uncommitted candidate $\widetilde G^{(t+1)}$. If a group introduces a structural conflict, only that group is discarded and merging continues. If no effective edit remains, the graph and reference are retained. After a full-pool rollout, let $n_{\mathrm{corrected}}^{(t)}$ and $n_{\mathrm{regressed}}^{(t)}$ count incorrect-to-correct and correct-to-incorrect transitions relative to $G^{(t)}$. Then
\[
G^{(t+1)}=
\begin{cases}
\widetilde G^{(t+1)}, &
n_{\mathrm{corrected}}^{(t)}>n_{\mathrm{regressed}}^{(t)},\\
G^{(t)}, & \text{otherwise}.
\end{cases}
\]
Positive net gain commits the candidate and its rollout as the next reference; otherwise, both graph and reference are restored.

Thus, one graph connects semantic activation, targeted updating, validation, and rollback.

\section{Experiments}
\label{sec:experiments}

\subsection{Experimental Setup}

\paragraph{Evaluation objectives.}
The experiments address three questions: how much does converting text skills into a graph and optimizing it improve performance, how does the method compare with existing skill optimizers, and what roles do node fields, edges, attribution, and the two Gates play? Further controlled comparisons test the choice of edit targets. Coverage analyses examine nodes and relations omitted by Top-6 retrieval and candidate regressions outside local validation. Additional evaluations report optimization cost, sequential interaction, and epoch-level commitments.

\paragraph{Models.}
We evaluate OpenAI's GPT-5.6-sol, GPT-5.5, and GPT-5.4-mini~\citep{openai2026models}; Google's Gemini-3.5-Flash~\citep{google2026gemini35flashdoc}; and Alibaba's Qwen3.8-Flash~\citep{alibaba2026qwen38flashdoc} and Qwen3.6-35B-A3B~\citep{qwen2026github}.

\paragraph{Benchmarks.}
We use three main benchmarks spanning open-domain question answering, mathematical reasoning, and document understanding:
\begin{enumerate}
\item \textbf{SearchQA}~\citep{dunn2017searchqa}: open-domain question answering from retrieved evidence.
\item \textbf{LiveMathematicianBench}~\citep{he2026livemathematicianbench}, abbreviated LiveMath: research-level multi-step mathematical reasoning.
\item \textbf{DocVQA}~\citep{mathew2021docvqa}: document-image question answering combining visual localization and text understanding.
\end{enumerate}
A separate ALFWorld experiment~\citep{shridhar2020alfworld} evaluates sequential interaction.

\paragraph{Evaluation metrics and repeated runs.}
SearchQA and LiveMath use exact-match accuracy; DocVQA is correct only when ANLS$=1$. Unless noted otherwise, we report the mean and half-range, $(\max-\min)/2$, over three runs. For graph-optimization settings, these are independent optimization-and-evaluation runs with seeds 42, 43, and 44; for frozen settings, the skill remains unchanged and only evaluation is repeated. Full settings are in Appendix~\ref{app:protocol}.

\subsection{Main Results and Progressive Analysis}

Table~\ref{tab:table-a} compares five successive settings under the same data splits and frozen model parameters. \emph{No skill} uses no external skill. \emph{Flat skill} directly uses the original SkillOpt text. \emph{Structured skill G0} uses an LLM to convert the same source skill into an initial graph, which is then held fixed for evaluation. \emph{G0 + Trace/Usage} additionally asks the model to record its execution path and the nodes and edges used. \emph{GraphSkillAA} updates the graph from success--failure contrasts and checks the edits with the Local and Big Gates.

\begin{table*}[t]
\centering
\renewcommand{\arraystretch}{0.96}
\setlength{\tabcolsep}{1.0mm}
\footnotesize
\resizebox{0.95\textwidth}{!}{%
\begin{tabular}{llccccc}
\toprule
\rowcolor{tableheadergray}
\textbf{Model} & \textbf{Benchmark} & \textbf{No skill} & \textbf{Flat skill} & \makecell[c]{\textbf{Structured}\\\textbf{skill G0}} & \makecell[c]{\textbf{G0 + Trace/}\\\textbf{Usage}} & \textbf{GraphSkillAA} \\
\midrule
\multirow{3}{*}{GPT-5.6-sol}
 & SearchQA & \hrange{66.0}{2.5} & \hrange{72.5}{0.5} & \hrange{75.0}{1.8} & \hrange{76.2}{1.5} & \textbf{\hrange{81.5}{1.3}} \\
 & LiveMath & \hrange{28.2}{1.7} & \hrange{42.2}{3.0} & \hrange{44.2}{2.6} & \hrange{51.3}{3.4} & \textbf{\hrange{66.7}{2.1}} \\
 & DocVQA & \hrange{68.0}{3.3} & \hrange{87.7}{0.5} & \hrange{88.0}{1.3} & \hrange{88.2}{2.0} & \textbf{\hrange{91.2}{1.0}} \\
\midrule
\multirow{3}{*}{Gemini-3.5-Flash}
 & SearchQA & \hrange{75.0}{0.5} & \hrange{77.5}{0.5} & \hrange{77.5}{0.5} & \hrange{77.5}{1.0} & \textbf{\hrange{79.0}{0.5}} \\
 & LiveMath & \hrange{34.5}{2.1} & \hrange{49.6}{2.1} & \hrange{52.4}{0.4} & \hrange{53.3}{3.4} & \textbf{\hrange{62.1}{2.6}} \\
 & DocVQA & \hrange{92.8}{0.5} & \hrange{94.0}{1.0} & \hrange{93.8}{0.5} & \hrange{93.5}{0.5} & \textbf{\hrange{95.0}{0.5}} \\
\midrule
\multirow{3}{*}{Qwen3.8-Flash}
 & SearchQA & \hrange{70.7}{1.3} & \hrange{74.2}{1.8} & \hrange{74.5}{1.5} & \hrange{72.2}{0.8} & \textbf{\hrange{76.5}{0.8}} \\
 & LiveMath & \hrange{30.2}{2.1} & \hrange{31.6}{4.3} & \hrange{56.7}{0.4} & \hrange{55.3}{3.4} & \textbf{\hrange{60.1}{3.4}} \\
 & DocVQA & \hrange{93.5}{0.5} & \hrange{92.8}{0.5} & \hrange{94.2}{0.8} & \hrange{93.3}{0.3} & \textbf{\hrange{95.0}{0.8}} \\
\bottomrule
\end{tabular}%
}
\caption{Progressive skill-graph construction and optimization (\%).}
\label{tab:table-a}

\vspace{6pt}
\renewcommand{\arraystretch}{0.96}
\setlength{\tabcolsep}{0.9mm}
\footnotesize
\begin{tabular*}{0.95\textwidth}{@{\extracolsep{\fill}}llcccccc@{}}
\toprule
\rowcolor{tableheadergray}
\textbf{Benchmark} & \textbf{Model} & \textbf{No skill} & \textbf{Trace2Skill} & \textbf{TextGrad} & \textbf{GEPA} & \textbf{SkillOpt} & \textbf{GraphSkillAA} \\
\midrule
\multirow{3}{*}{SearchQA}
 & GPT-5.5 & 77.7 & 82.4 & 81.4 & 84.8 & 82.6 & \textbf{\hrange{87.0}{0.6}} \\
 & GPT-5.4-mini & 75.9 & 78.6 & 77.5 & 79.4 & 81.3 & \textbf{\hrange{82.6}{1.0}} \\
 & Qwen3.6-35B-A3B & 72.7 & 75.4 & 76.4 & 75.8 & 80.7 & \textbf{\hrange{83.4}{0.5}} \\
\midrule
\multirow{3}{*}{LiveMath}
 & GPT-5.5 & 37.6 & 51.9 & 49.2 & 43.3 & 51.1 & \textbf{\hrange{58.6}{2.4}} \\
 & GPT-5.4-mini & 14.8 & 32.8 & 27.2 & 27.2 & 32.8 & \textbf{\hrange{36.3}{2.0}} \\
 & Qwen3.6-35B-A3B & 31.2 & 29.6 & 7.3 & 31.2 & 28.0 & \textbf{\hrange{51.9}{1.2}} \\
\midrule
\multirow{3}{*}{DocVQA}
 & GPT-5.5 & 78.8 & 90.6 & 87.2 & 89.1 & 89.7 & \textbf{\hrange{91.1}{0.7}} \\
 & GPT-5.4-mini & 71.4 & 88.5 & 84.0 & 83.7 & 82.2 & \textbf{\hrange{89.0}{3.2}} \\
 & Qwen3.6-35B-A3B & 87.6 & 90.4 & 84.5 & 88.0 & 91.4 & \textbf{\hrange{92.7}{1.1}} \\
\bottomrule
\end{tabular*}
\caption{Comparison with skill-evolution and text-optimization methods under the SkillOpt direct-chat protocol (\%).}
\label{tab:table-b}
\end{table*}

\GraphSkillAA{} achieves the highest mean for every model--task combination. Flat skill improves over No skill in most settings, indicating that the initial skills already provide useful task procedures. For GPT-5.6-sol, converting Flat skill into Structured skill G0 improves SearchQA, LiveMath, and DocVQA by $2.5$, $2.0$, and $0.3\pp$, respectively. Adding Trace/Usage to the fixed G0 yields further gains of $1.2$, $7.1$, and $0.2\pp$. Updating and validating the graph brings further gains of $5.3$, $15.4$, and $3.0\pp$ over G0 + Trace/Usage, for total improvements of $9.0$, $24.5$, and $3.5\pp$ over Flat skill. Other models vary in their intermediate-stage changes, but the complete method outperforms G0 + Trace/Usage in all nine settings. Recording execution alone therefore does not reproduce the gains from graph optimization.

Under the GPT-5.6-sol main protocol, \GraphSkillAA{} uses $2.12$, $3.56$, and $3.87$ times SkillOpt's actual optimization tokens on SearchQA, LiveMath, and DocVQA, respectively. Accuracy improves by $9.0$, $24.5$, and $3.5\pp$ over Flat skill, which directly uses SkillOpt's best skills. Table~\ref{tab:optimization-cost} in Appendix~\ref{app:model-settings} gives the cost--performance comparison.

\subsection{Comparison with Existing Methods}

Table~\ref{tab:table-b} uses GPT-5.5, GPT-5.4-mini~\citep{openai2026models}, and Qwen3.6-35B-A3B~\citep{qwen2026github} to compare \GraphSkillAA{} with No skill, Trace2Skill~\citep{trace2skill2026}, TextGrad~\citep{yuksekgonul2024textgrad}, GEPA~\citep{agrawal2025gepa}, and SkillOpt~\citep{skillopt2026}. The comparison follows the direct-chat protocol of SkillOpt Table~1, using the same released instances, test IDs, original inputs, inference budget, scorer, and approximately $2{:}1{:}7$ train/validation/test split. \GraphSkillAA{} updates only on train/validation and reserves test for final evaluation. The GPT-5.5 SkillOpt results in Table~\ref{tab:table-b} are means of three local reevaluations of the originally released best skills, using the original data splits, test examples, and scoring method. Exact split and protocol details are in Appendix~\ref{app:protocol}.

The baselines use different optimization strategies. Trace2Skill derives skills from execution trajectories, TextGrad guides edits with textual feedback, GEPA revises prompts from complete trajectories, and SkillOpt improves skill documents through bounded edits and validation. \GraphSkillAA{} instead specifies edits to individual node fields and relations and checks the effects of both repair groups and the merged graph.

GraphSkillAA ranks first in all nine model--task combinations. It exceeds the strongest alternative in every row by $1.3$--$2.7\pp$ on SearchQA, $3.5$--$20.7\pp$ on LiveMath, and $0.5$--$1.3\pp$ on DocVQA. The largest margins occur on LiveMath, where the complete optimization workflow is particularly beneficial for multi-step reasoning.

\subsection{Component Ablations and Mechanism Analysis}

Table~\ref{tab:table-c} examines graph structure, retrieval, updating, and teacher choice, using the same metrics as Table~\ref{tab:table-a}. Graph-structure ablations test nodes and edges. Retrieval ablations keep the optimized graph fixed and change only the content shown to the student. Update ablations remove attribution, successful-case protection, the Local Gate, or the Big Gate individually. The final two rows report GraphSkillAA (Full) under two teacher settings, with student, data, and the three-epoch schedule fixed. Component ablations use self-teachers, and six-setting mean gains compare each student with its own self-teacher Full. Colored superscripts show differences from Flat skill. Appendix~\ref{app:ablations} gives the detailed settings.

\begin{table*}[t]
\centering
\renewcommand{\arraystretch}{1.04}
\setlength{\tabcolsep}{0.9mm}
\scriptsize
\resizebox{\textwidth}{!}{%
\begin{tabular}{llllllll}
\toprule
\multirow{2}{*}{\textbf{Category}} & \multirow{2}{*}{\textbf{Component}}
& \multicolumn{3}{>{\columncolor{tableheadergray}}c}{\textbf{Student: GPT-5.6-sol}}
& \multicolumn{3}{>{\columncolor{tableheadergray}}c}{\textbf{Student: GPT-5.4-mini}} \\
\cmidrule(lr){3-5}\cmidrule(lr){6-8}
& & \multicolumn{1}{>{\columncolor{tableheadergray}}c}{\textbf{SearchQA}}
& \multicolumn{1}{>{\columncolor{tableheadergray}}c}{\textbf{LiveMath}}
& \multicolumn{1}{>{\columncolor{tableheadergray}}c}{\textbf{DocVQA}}
& \multicolumn{1}{>{\columncolor{tableheadergray}}c}{\textbf{SearchQA}}
& \multicolumn{1}{>{\columncolor{tableheadergray}}c}{\textbf{LiveMath}}
& \multicolumn{1}{>{\columncolor{tableheadergray}}c}{\textbf{DocVQA}} \\
\midrule
\multirow[c]{2}{*}{\makecell[l]{Graph structure\\ablation}}
& Flat skill
& \hrange{72.5}{0.5} & \hrange{44.2}{2.6} & \hrange{85.5}{1.3}
& \hrange{74.2}{1.0} & \hrange{24.8}{2.1} & \hrange{67.5}{15.5} \\
& Nodes only
& \hrange{77.5}{0.5}\gain{5.0} & \hrange{62.1}{2.1}\gain{17.9} & \hrange{90.0}{1.0}\gain{4.5}
& \hrange{77.0}{0.8}\gain{2.8} & \hrange{21.4}{0.0}\drop{3.4} & \hrange{86.5}{1.8}\gain{19.0} \\
\midrule
\multirow[c]{4}{*}{\makecell[l]{Retrieval mechanism\\ablation}}
& No when-to-use \& avoid
& \hrange{70.5}{1.8}\drop{2.0} & \hrange{42.7}{4.3}\drop{1.5} & \hrange{84.3}{1.0}\drop{1.2}
& \hrange{73.5}{1.5}\drop{0.7} & \hrange{19.9}{0.4}\drop{4.9} & \hrange{78.2}{1.8}\gain{10.7} \\
& No edge-rationale
& \hrange{78.5}{0.8}\gain{6.0} & \hrange{63.5}{6.0}\gain{19.3} & \hrange{89.0}{0.5}\gain{3.5}
& \hrange{76.5}{0.5}\gain{2.3} & \hrange{26.5}{0.9}\gain{1.7} & \hrange{88.0}{0.5}\gain{20.5} \\
& S--Q retrieval
& \hrange{73.5}{2.3}\gain{1.0} & \hrange{42.5}{1.7}\drop{1.7} & \hrange{86.0}{0.8}\gain{0.5}
& \hrange{72.5}{1.0}\drop{1.7} & \hrange{21.1}{2.6}\drop{3.7} & \hrange{82.8}{0.5}\gain{15.3} \\
& Nearest-Q reuse
& \hrange{77.7}{1.3}\gain{5.2} & \hrange{58.1}{1.7}\gain{13.9} & \hrange{88.5}{1.0}\gain{3.0}
& \hrange{76.5}{1.5}\gain{2.3} & \hrange{25.4}{3.4}\gain{0.6} & \hrange{87.7}{0.8}\gain{20.2} \\
\midrule
\multirow[c]{4}{*}{\makecell[l]{Update mechanism\\ablation}}
& No structured attribution
& \hrange{73.2}{1.5}\gain{0.7} & \hrange{38.5}{5.6}\drop{5.7} & \hrange{86.7}{4.3}\gain{1.2}
& \hrange{73.0}{1.0}\drop{1.2} & \hrange{18.5}{1.7}\drop{6.3} & \hrange{85.7}{2.5}\gain{18.2} \\
& No correct-case protection
& \hrange{80.5}{0.5}\gain{8.0} & \hrange{62.4}{3.8}\gain{18.2} & \hrange{89.2}{0.5}\gain{3.7}
& \hrange{75.5}{1.3}\gain{1.3} & \hrange{22.5}{0.4}\drop{2.3} & \hrange{87.3}{0.8}\gain{19.8} \\
& No Local Gate
& \hrange{78.0}{2.3}\gain{5.5} & \hrange{49.0}{5.6}\gain{4.8} & \hrange{89.5}{1.8}\gain{4.0}
& \hrange{74.7}{2.3}\gain{0.5} & \hrange{20.2}{3.4}\drop{4.6} & \hrange{84.5}{2.8}\gain{17.0} \\
& No Big Gate
& \hrange{80.0}{1.3}\gain{7.5} & \hrange{62.1}{11.1}\gain{17.9} & \hrange{90.0}{1.0}\gain{4.5}
& \hrange{76.8}{0.8}\gain{2.6} & \hrange{25.9}{1.3}\gain{1.1} & \hrange{87.2}{0.5}\gain{19.7} \\
\midrule
\multirow[c]{2}{*}{\makecell[l]{GraphSkillAA (Full)\\teacher comparison}}
& GPT-5.4-mini teacher
& \hrange{78.2}{0.5}\gain{5.7} & \hrange{64.7}{1.7}\gain{20.5} & \hrange{90.0}{1.3}\gain{4.5}
& \textbf{\hrange{75.7}{0.5}}\gain{1.5} & \textbf{\hrange{23.9}{3.0}}\drop{0.9} & \textbf{\hrange{86.3}{1.3}}\gain{18.8} \\
& GPT-5.6-sol teacher
& \textbf{\hrange{81.5}{1.3}}\gain{9.0} & \textbf{\hrange{66.7}{2.1}}\gain{22.5} & \textbf{\hrange{91.2}{1.0}}\gain{5.7}
& \hrange{78.2}{0.8}\gain{4.0} & \hrange{29.1}{2.1}\gain{4.3} & \hrange{89.0}{1.0}\gain{21.5} \\
\bottomrule
\end{tabular}%
}
\caption{Component ablations and mechanism analysis (\%). Column groups identify students; the final two rows identify teachers. Bold cells mark each student's self-teacher GraphSkillAA (Full) reference for component ablations. Six-setting mean gains use these six bold cells. Colored superscripts show differences from Flat skill (red: increase; green: decrease). Teacher transfer compares the two teacher rows within each student.}
\label{tab:table-c}
\end{table*}

\paragraph{Graph-structure ablation.}\label{sec:graph-evidence}
Relative to \GraphSkillAA{} (Full) with the same model serving as teacher and student, removing all edges (Nodes only) lowers performance by $1.8\pp$ on average. GPT-5.6-sol drops by $4.0$, $4.6$, and $1.2\pp$ on the three tasks; retaining edges but hiding their explanations lowers accuracy by $3.0$, $3.2$, and $2.2\pp$. Relations and their explanations thus support skill composition with unchanged node content. Changes for GPT-5.4-mini vary by task. Even without edges, the optimized nodes remain $7.6\pp$ above Flat skill on average, combining gains from structured representation and node-content optimization.

\paragraph{Retrieval-mechanism ablation.}
Keeping the optimized graph fixed and hiding only \texttt{when\_to\_use} and \texttt{avoid} lowers all six results by $9.4\pp$ on average, indicating that applicability and exclusion conditions help the model select and use rules. S--Q retrieval retains only the top six nodes and their internal relations and scores $7.8\pp$ below the same student's self-teacher \GraphSkillAA{} (Full) on average. The coverage analysis in Appendix~\ref{app:retrieval-coverage} reports node and edge coverage of 69.8\% and 56.4\% against Full's usage records on LiveMath, showing that retrieval omits both rules and their connections. Nearest-Q reuse and removing edge rationales both score below the self-teacher complete graph for GPT-5.6-sol but above it for GPT-5.4-mini, reflecting differences in how the models use graph information.

\paragraph{Update-mechanism ablation.}
Replacing structured attribution with generic failure summaries lowers all six results by $8.3\pp$ on average. This control uses the same evidence-construction rules and retains the teacher, joint synthesis, and both Gates, indicating that structured attribution guides effective repairs. Removing the Local Gate also lowers all six results, by $4.9\pp$ on average, supporting group-wise retesting and rollback. Correct-case protection contributes $1.3\pp$ on average. The Big Gate uses update-pool net gains to decide whether to commit merged candidates. It improves the three GPT-5.6-sol results by $1.5$, $4.6$, and $1.2\pp$, with no test-set gain for GPT-5.4-mini.

\paragraph{Teacher-transfer evaluation.}
GPT-5.6-sol as teacher yields higher results in all six settings. Graphs produced by GPT-5.4-mini also improve the GPT-5.6-sol student over Flat skill, showing that the teacher and student can use different models. Appendices~\ref{app:teacher-ablation}--\ref{app:matched-ablation} report the four pairings and their gains.

\subsection{Additional Evaluations and Analysis}
\label{sec:additional-analysis}

\paragraph{Attribution-target comparison.}
With models, evidence-construction rules, patch budgets, and Gate settings fixed, we randomly replace diagnosed targets with other type-compatible graph objects. Original attribution achieves higher net repair gains on all three tasks and improves test accuracy by $3.2$, $6.0$, and $1.5\pp$, showing that choosing appropriate edit targets improves repair outcomes (Appendix~\ref{app:shuffled-attribution}).

\paragraph{ALFWorld evaluation.}
SkillOpt Markdown and Initial Graph skill both achieve 100\% success with a 100-step budget, while the initial graph scores $5.9\pp$ higher at 50 steps (Appendix~\ref{app:alfworld}). These results show how skill representation and execution budgets affect task completion. \GraphSkillAA{} uses failure evidence to decide whether skills should change and returns \texttt{NO\_PATCH} when no reusable defect is identified, preserving effective procedures.

\paragraph{Epoch-level commit behavior.}
The three-epoch trajectories show that the Gates reject candidates that fail the acceptance criteria and preserve the existing graph and its reference results (Appendix~\ref{app:optimization-dynamics}). The Big-Gate subset evaluation in Appendix~\ref{app:local-gate-details} finds outside-scope regression rates of 3.02\%--9.86\% for merged candidates, supporting whole-graph checks after local retesting.

\section{Conclusion}

\GraphSkillAA{} organizes skills in a graph and uses execution feedback to revise specific rules and relations without updating model parameters. The teacher identifies repair targets through success--failure contrasts, the Local Gate checks each repair group, and the Big Gate decides whether to commit the merged graph. Rejected edits can be reverted by group or by epoch. Across three benchmarks and multiple model families, the method achieves the highest mean in every main setting and ranks first in all nine method comparisons. Progressive experiments and ablations show gains from explicit applicability and composition relations, structured attribution, and group-wise retesting. Graph structure thus connects skill use, repair, and validation. Future work can explore combining external skill updates with model training.

\phantomsection\label{main:end}
\clearpage
\bibliography{graphopt_paper_latex}
\bibliographystyle{iclr2027_conference}

\clearpage
\appendix
\onecolumn

\section*{\centering \LARGE Appendix Contents}
\addcontentsline{toc}{section}{Appendix Contents}

\vspace{0.45cm}
\begingroup
\setlength{\parindent}{0pt}
\setlength{\parskip}{0.10em}
\fontsize{10.5pt}{14pt}\selectfont

\noindent\hyperref[app:reproducibility]{\textbf{A \quad Reproducibility Statement}}\dotfill\pageref{app:reproducibility}\par

\vspace{0.55em}

\noindent\hyperref[app:ai-use]{\textbf{B \quad AI Use Statement}}\dotfill\pageref{app:ai-use}\par

\vspace{0.55em}

\noindent\hyperref[app:implementation]{\textbf{C \quad Additional Evaluations and Method Specification}}\dotfill\pageref{app:implementation}\par
\noindent\hspace*{1.5em}\hyperref[app:alfworld]{C.1. \quad Sequential Interaction under Step Budgets}\dotfill\pageref{app:alfworld}\par
\noindent\hspace*{1.5em}\hyperref[app:optimization-dynamics]{C.2. \quad Epoch-Level Commit Dynamics}\dotfill\pageref{app:optimization-dynamics}\par
\noindent\hspace*{1.5em}\hyperref[app:skillaa-overview]{C.3. \quad End-to-End GraphSkillAA Pipeline}\dotfill\pageref{app:skillaa-overview}\par
\noindent\hspace*{1.5em}\hyperref[app:update-contract]{C.4. \quad Epoch-Level Execution and Usage Records}\dotfill\pageref{app:update-contract}\par
\noindent\hspace*{1.5em}\hyperref[app:legal-edits]{C.5. \quad Allowed Graph Edits and Atomic Groups}\dotfill\pageref{app:legal-edits}\par
\noindent\hspace*{1.5em}\hyperref[app:edit-routing]{C.6. \quad Attribution Routing and Case Studies}\dotfill\pageref{app:edit-routing}\par
\noindent\hspace*{1.5em}\hyperref[app:local-gate-details]{C.7. \quad Gate Scope Expansion and Rollback Mechanism}\dotfill\pageref{app:local-gate-details}\par
\noindent\hspace*{1.5em}\hyperref[app:validation-levels]{C.8. \quad Validation Checks and Outcomes}\dotfill\pageref{app:validation-levels}\par

\vspace{0.55em}

\noindent\hyperref[app:prompts]{\textbf{D \quad Prompt Templates}}\dotfill\pageref{app:prompts}\par
\noindent\hspace*{1.5em}\hyperref[app:searchqa-graph-workflow]{D.1. \quad From Initial Skills to Graphs: Conversion and Execution}\dotfill\pageref{app:searchqa-graph-workflow}\par
\noindent\hspace*{1.5em}\hyperref[app:skillgraph-example]{D.2. \quad Abbreviated SearchQA SkillGraph Example}\dotfill\pageref{app:skillgraph-example}\par
\noindent\hspace*{1.5em}\hyperref[app:prompt-student]{D.3. \quad Student Execution and Trace/Usage Specification}\dotfill\pageref{app:prompt-student}\par
\noindent\hspace*{1.5em}\hyperref[app:prompt-attribution]{D.4. \quad Abductive Attribution Prompt}\dotfill\pageref{app:prompt-attribution}\par
\noindent\hspace*{1.5em}\hyperref[app:prompt-meta]{D.5. \quad Epoch-Level Meta-Auditor Prompt}\dotfill\pageref{app:prompt-meta}\par
\noindent\hspace*{1.5em}\hyperref[app:prompt-patch-gate]{D.6. \quad Patch Synthesis and Gate Audit Prompts}\dotfill\pageref{app:prompt-patch-gate}\par

\vspace{0.55em}

\noindent\hyperref[app:protocol]{\textbf{E \quad Experimental Protocol}}\dotfill\pageref{app:protocol}\par
\noindent\hspace*{1.5em}\hyperref[app:benchmark-splits]{E.1. \quad Dataset Splits and Group Construction}\dotfill\pageref{app:benchmark-splits}\par
\noindent\hspace*{1.5em}\hyperref[app:model-settings]{E.2. \quad Model Access, Inference, and Optimization Settings}\dotfill\pageref{app:model-settings}\par
\noindent\hspace*{1.5em}\hyperref[app:artifact-provenance]{E.3. \quad Evaluation Protocol}\dotfill\pageref{app:artifact-provenance}\par
\noindent\hspace*{1.5em}\hyperref[app:comparison-ledger]{E.4. \quad Result Sources and Experimental Settings}\dotfill\pageref{app:comparison-ledger}\par

\vspace{0.55em}

\noindent\hyperref[app:ablations]{\textbf{F \quad Ablation Protocols and Retrieval Definitions}}\dotfill\pageref{app:ablations}\par
\noindent\hspace*{1.5em}\hyperref[app:graph-ablation]{F.1. \quad Shared Settings and Graph-Structure Ablation}\dotfill\pageref{app:graph-ablation}\par
\noindent\hspace*{1.5em}\hyperref[app:retrieval]{F.2. \quad Retrieval-Mechanism Ablations}\dotfill\pageref{app:retrieval}\par
\noindent\hspace*{1.5em}\hyperref[app:update-ablation]{F.3. \quad Update-Mechanism Ablations}\dotfill\pageref{app:update-ablation}\par
\noindent\hspace*{1.5em}\hyperref[app:teacher-ablation]{F.4. \quad Teacher-Transfer Evaluation}\dotfill\pageref{app:teacher-ablation}\par
\noindent\hspace*{1.5em}\hyperref[app:matched-ablation]{F.5. \quad Comparison Protocol for Table~\ref{tab:table-c}}\dotfill\pageref{app:matched-ablation}\par
\noindent\hspace*{1.5em}\hyperref[app:shuffled-attribution]{F.6. \quad Shuffling Initial Edit Targets}\dotfill\pageref{app:shuffled-attribution}\par
\noindent\hspace*{1.5em}\hyperref[app:retrieval-coverage]{F.7. \quad Retrieval Coverage of Used Nodes and Relations}\dotfill\pageref{app:retrieval-coverage}\par

\vspace{0.55em}

\noindent\hyperref[app:badcases]{\textbf{G \quad Residual Error Analysis}}\dotfill\pageref{app:badcases}\par

\vspace{0.55em}

\noindent\hyperref[app:discussion]{\textbf{H \quad Discussion, Limitations, and Future Work}}\dotfill\pageref{app:discussion}\par
\noindent\hspace*{1.5em}\hyperref[app:interpretation]{H.1. \quad Interpretation of Experimental Findings}\dotfill\pageref{app:interpretation}\par
\noindent\hspace*{1.5em}\hyperref[sec:limitations]{H.2. \quad Limitations}\dotfill\pageref{sec:limitations}\par
\noindent\hspace*{1.5em}\hyperref[app:research-directions]{H.3. \quad Long-Term Research Directions}\dotfill\pageref{app:research-directions}\par
\endgroup
\vspace{0.8cm}
\clearpage

\section{Reproducibility Statement}
\label{app:reproducibility}

The supplementary material describes the graph schema, edit rules, attribution and Gate procedures, data splits, model settings, and scorers. We will release the code, experiment configurations, data splits, final skill graphs, and per-example predictions.

\section{AI Use Statement}
\label{app:ai-use}

Generative AI tools assisted with language editing, code-level consistency checks, literature search, and LaTeX conversion. The authors are responsible for the method, experimental design, result verification, literature selection, and the manuscript as a whole.

\section{Additional Evaluations and Method Specification}
\label{app:implementation}
\label{app:additional-experiments}

\subsection{Sequential Interaction under Step Budgets}
\label{app:alfworld}

ALFWorld is used to examine how the step budget affects task completion. Table~\ref{tab:table-d} follows the SkillOpt ALFWorld protocol independently of the three main benchmarks.

\begin{table*}[t]
\centering
\renewcommand{\arraystretch}{0.96}
\setlength{\tabcolsep}{2.4mm}
\footnotesize
\resizebox{0.75079\textwidth}{!}{%
\begin{tabular}{lccccc}
\toprule
\rowcolor{tableheadergray}
\textbf{Setting} & \textbf{S@50} & \textbf{S@100} & \textbf{$\Delta$S} & \textbf{Steps@50} & \textbf{TO@50} \\
\midrule
SkillOpt Markdown & \hrange{89.6}{1.1} & \textbf{\hrange{100.0}{0.0}} & +10.4 & \hrange{9.7}{0.1} & \hrange{10.4}{1.1} \\
Initial Graph skill & \textbf{\hrange{95.5}{0.4}} & \textbf{\hrange{100.0}{0.0}} & \textbf{+4.5} & \textbf{\hrange{9.5}{0.2}} & \textbf{\hrange{4.5}{0.4}} \\
\bottomrule
\end{tabular}%
}
\caption{ALFWorld step-budget evaluation. S@50/S@100: success within 50/100 steps; $\Delta$S: absolute percentage-point gain; Steps@50: steps among successful runs; TO@50: timeout rate.}
\label{tab:table-d}
\end{table*}

Both settings reach 100\% success within 100 steps. Under the stricter 50-step budget, Initial Graph skill achieves a higher success rate, shorter successful paths, and fewer timeouts than SkillOpt Markdown. Tasks not completed within 50 steps are recorded as timeouts and all complete under the extended budget. During failure analysis, the system examines rule content and execution records for evidence of a reusable defect, retaining the original graph through \texttt{NO\_PATCH} when the evidence is insufficient.

\subsection{Epoch-Level Commit Dynamics}
\label{app:optimization-dynamics}

\begin{figure*}[t]
\centering
\includegraphics[width=0.96\textwidth]{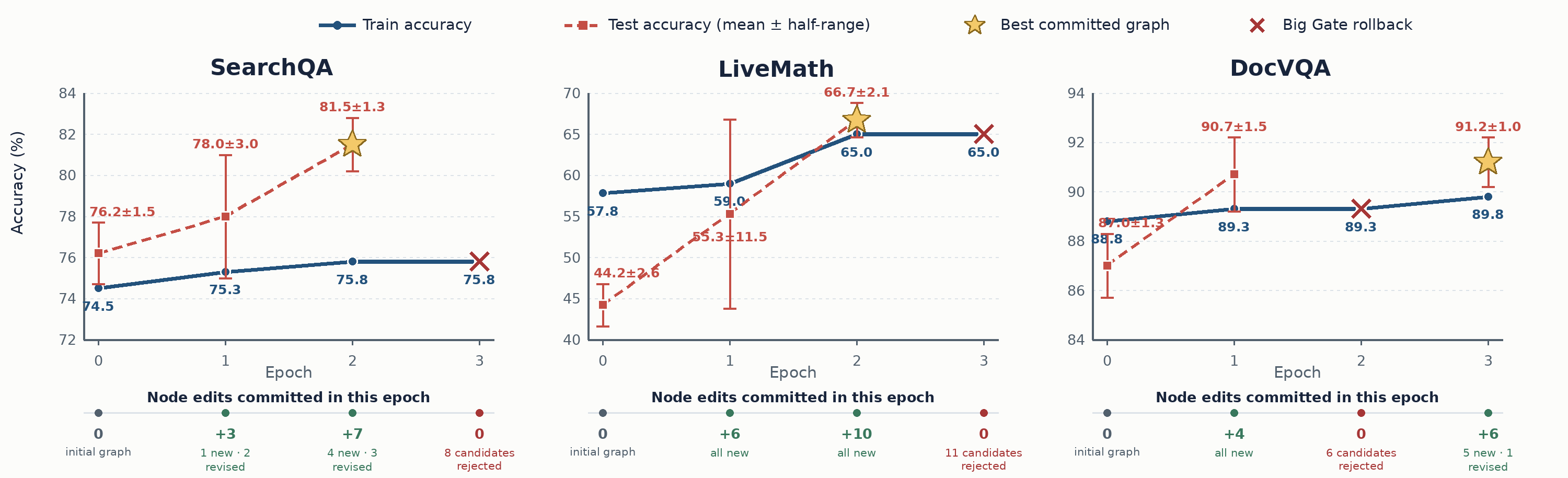}
\setlength{\abovecaptionskip}{3pt}
\setlength{\belowcaptionskip}{0pt}
\caption{GPT-5.6-sol GraphSkillAA epoch trajectories (\%). Blue annotations report initial and final committed-graph accuracy on the complete update pool, averaged across three runs. Dashed red lines show test means, with error bars denoting half-ranges over three runs. Epoch-zero test points use the G0 + Trace/Usage results from Table~\ref{tab:table-a} as the initial baseline. Gold stars mark the best committed graph. The lower strip reports committed node edits and marks Big-Gate rollback with red crosses. The test set is not reevaluated when the graph remains unchanged; node counts exclude edge edits.}
\label{fig:epoch-trajectories}
\end{figure*}

Figure~\ref{fig:epoch-trajectories} presents the GPT-5.6-sol runs summarized in Table~\ref{tab:table-a}, including whether each epoch's candidate passed validation.

SearchQA and LiveMath obtain their best graphs at epoch 2. Their epoch-3 candidates fail the Big Gate, so the previous graphs remain in use. DocVQA rolls back at epoch 2 and accepts new repairs at epoch 3.

\subsection{End-to-End GraphSkillAA Pipeline}
\label{app:skillaa-overview}

Each epoch starts from the current graph $G^{(t)}$ and its reference results $R$ on the complete update pool. If a candidate passes the Big Gate, the graph and reference results are updated together. Otherwise, both remain unchanged. Algorithm~\ref{alg:skillaa-overview} gives the complete procedure, Algorithm~\ref{alg:attribution} details attribution, and Appendix~\ref{app:local-gate-details} defines how examples are selected for retesting.

\begin{algorithm}[t]
\caption{Complete GraphSkillAA pipeline}
\label{alg:skillaa-overview}
\footnotesize
\begin{algorithmic}[1]
\Require frozen $M_{\mathrm{student}}$, frozen $M_{\mathrm{teacher}}$, update pool $\mathcal D_{\mathrm{update}}$, fixed update quartets $\mathcal Q$, initial graph $G^{(0)}$, epochs $E$
\Ensure final committed graph $G^{(E)}$
\State $R \gets \Call{Rollout}{M_{\mathrm{student}},G^{(0)},\mathcal D_{\mathrm{update}}}$
\For{$t=0,\ldots,E-1$}
    \State $F \gets$ failed update examples in $R$
    \State $D \gets \Call{Attribute}{M_{\mathrm{teacher}},F,G^{(t)},R,\mathcal Q}$ \Comment{Algorithm~\ref{alg:attribution}; successful siblings as contrasts}
    \State $P \gets \Call{TargetedPatch}{D,R,\mathcal Q}$ \Comment{initial route constraints and success guards}
    \State $P^\star \gets \Call{JointSemanticSynthesis}{P,D,R}$ \Comment{synthesize connected edits within localized targets}
    \State $\{J_k\}_{k=1}^{K} \gets \Call{AtomicGroups}{P^\star}$
    \State $S \gets \emptyset$
    \For{each $J_k$}
        \State $b\gets\Call{LocalGate}{G^{(t)},J_k,R}$ \Comment{Appendix~\ref{app:local-gate-details}}
        \If{$\neg b$ and $J_k$ is refinable}
            \State $J_k\gets\Call{RefineRejectedPatch}{J_k}$
            \State $b\gets\Call{LocalGate}{G^{(t)},J_k,R}$ \Comment{at most once}
        \EndIf
        \If{$b$}
            \State $S \gets S\cup\{J_k\}$
        \EndIf
    \EndFor
    \State $\widetilde G \gets \Call{ApplyAll}{G^{(t)},S}$ \Comment{merge in order; skip each conflicting group}
    \If{$\widetilde G=G^{(t)}$}
        \State $G^{(t+1)}\gets G^{(t)}$; keep $R$ unchanged; \textbf{continue}
    \EndIf
    \State $\widetilde R \gets \Call{Rollout}{M_{\mathrm{student}},\widetilde G,\mathcal D_{\mathrm{update}}}$
    \If{$\Call{NumCorrected}{R,\widetilde R}>\Call{NumRegressed}{R,\widetilde R}$}
        \State $G^{(t+1)}\gets\widetilde G$
        \State $R\gets\widetilde R$
    \Else
        \State $G^{(t+1)}\gets G^{(t)}$
        \State keep $R$ unchanged
    \EndIf
\EndFor
\State \Return $G^{(E)}$
\end{algorithmic}
\end{algorithm}

\subsection{Epoch-Level Execution and Usage Records}
\label{app:update-contract}

Optimization starts by holding the graph fixed and running the student on the complete update pool to obtain pre-edit reference results: 800 examples each for SearchQA and DocVQA, and 468 for LiveMath. Requests may run concurrently, with editing starting after all results return. Fixed groups provide successful contrasts and expand the retest scope. Later epochs reuse existing results when the graph, rendered prompt, sample set, and run status are unchanged. An accepted Big-Gate candidate supplies the merged graph's results, whereas rollback retains the epoch-start results. Otherwise, the student is run again.

\paragraph{\texttt{usage} validation.}
\texttt{usage} collects the node and edge IDs that the student reports using, with duplicates removed. Each ID should correspond to an input--operation--output record in \texttt{trace}. The program checks record provenance, valid IDs and edge endpoints, and agreement between the two ID sets. This checks record consistency, while subsequent retesting evaluates the effect of the rules. A missing or unknown ID, an ID-set mismatch, or an invalid relation marks usage as unknown and includes the example in retesting, so that potentially affected examples are not omitted.

\subsection{Allowed Graph Edits and Atomic Groups}
\label{app:legal-edits}

\GraphSkillAA{} applies updates to the node IDs, fields, and relation endpoints specified by the initial attribution, permitting three classes of graph operations:
\begin{enumerate}
\item revise the relevant part of a node's \texttt{when\_to\_use}, \texttt{how\_to\_use}, or \texttt{avoid}, including replacement of a localized harmful rule;
\item add a complete specialist node with a supporting edge, or add or correct a \textsc{prereq}/\textsc{enhance} edge, including its direction, type, or strength; and
\item delete a node or edge only when deletion is enabled and supported by evidence. Fixed \texttt{co\_occur} edges and their endpoint nodes are excluded from deletion.
\end{enumerate}
Dependent operations are handled together. For example, a new node and its connecting edge, or an old edge and its replacement, are retained or rolled back as a unit. Existing titles and categories normally remain fixed. Insufficient evidence and execution lapses produce \texttt{NO\_PATCH} at the attribution stage.

\paragraph{Joint semantic synthesis.}
Initial attribution lists all edit targets, including necessary dependent operations. For example, a specialist node and its connecting edge are included in the same proposal. Joint synthesis organizes these proposals by connected nodes and edges, using failure evidence and related successes to coordinate edits within the localized fields or relations without expanding their scope. Each edit retains its source evidence. The resulting repair groups undergo structural checks and both Gates.

\paragraph{Merging and structural validation.}
Candidate edits are processed in a fixed order. Exact duplicates are removed, and edits that share evidence or touch the same node form one atomic group to prevent overwriting. Each group is first applied to a copy of the graph. The entire group is rejected if it creates a duplicate relation, conflicting \textsc{prereq}/\textsc{enhance} types on one directed pair, a missing endpoint, a self-loop, or a \textsc{prereq} cycle. Groups that pass the Local Gate are merged in proposal order. If adding a group creates a structural conflict, only that group is discarded, previously merged groups are retained, and processing continues with the remaining groups. The final check verifies the candidate's structure and its agreement with the graph reconstructed from the edit record before passing it to the Big Gate. If no effective edits remain, the original graph is retained.

\paragraph{Fixed co-occurrence relations.}
Teachers may edit \textsc{prereq} and \textsc{enhance} relations. \textsc{Co-occur} retains its initial settings from $G^{(0)}$ throughout all three optimization epochs. All edges use a source--target storage format, but the arrow on a co-occurrence edge implies neither execution order nor causality. Edge strength is either \texttt{medium} or \texttt{strong}, and only \textsc{prereq} requires the prerequisite node to be completed first.

\subsection{Attribution Routing and Case Studies}
\label{app:edit-routing}

Attribution routing uses the failure diagnosis to identify the node fields or edges to edit.

\begin{algorithm}[t]
\caption{Abductive attribution and edit routing}
\label{alg:attribution}
\footnotesize
\begin{algorithmic}[1]
\Require failure, current graph, verifier feedback, validated skill-use record, successful siblings
\Ensure root cause and an edit target or \texttt{NO\_PATCH}
\State Locate the earliest observable error and compare relevant differences with successful siblings
\If{the error or the difference cannot be reliably localized} \State \Return \texttt{INSUFFICIENT\_EVIDENCE} \EndIf
\State $C\gets$ graph nodes covering the required behavior
\If{$C=\varnothing$} \State \Return \texttt{MISSING\_SKILL\_FAMILY} \EndIf
\If{a correct node exists but was not used, and evidence supports a missing activation cue} \State \Return \texttt{MISSING\_ACTIVATION\_CUE} \EndIf
\If{stored semantics of a used object are harmful} \State \Return \texttt{HARMFUL\_EXISTING\_RULE} \EndIf
\If{a used node lacks a step or boundary, or a required relation is missing} \State \Return \texttt{MISSING\_OR\_INCORRECT\_PROCEDURE} \EndIf
\If{the graph is correct but execution diverges} \State \Return \texttt{EXECUTION\_LAPSE} \EndIf
\State \Return \texttt{INSUFFICIENT\_EVIDENCE}
\end{algorithmic}
\end{algorithm}

\begin{figure*}[t]
\centering
\includegraphics[width=0.96\textwidth]{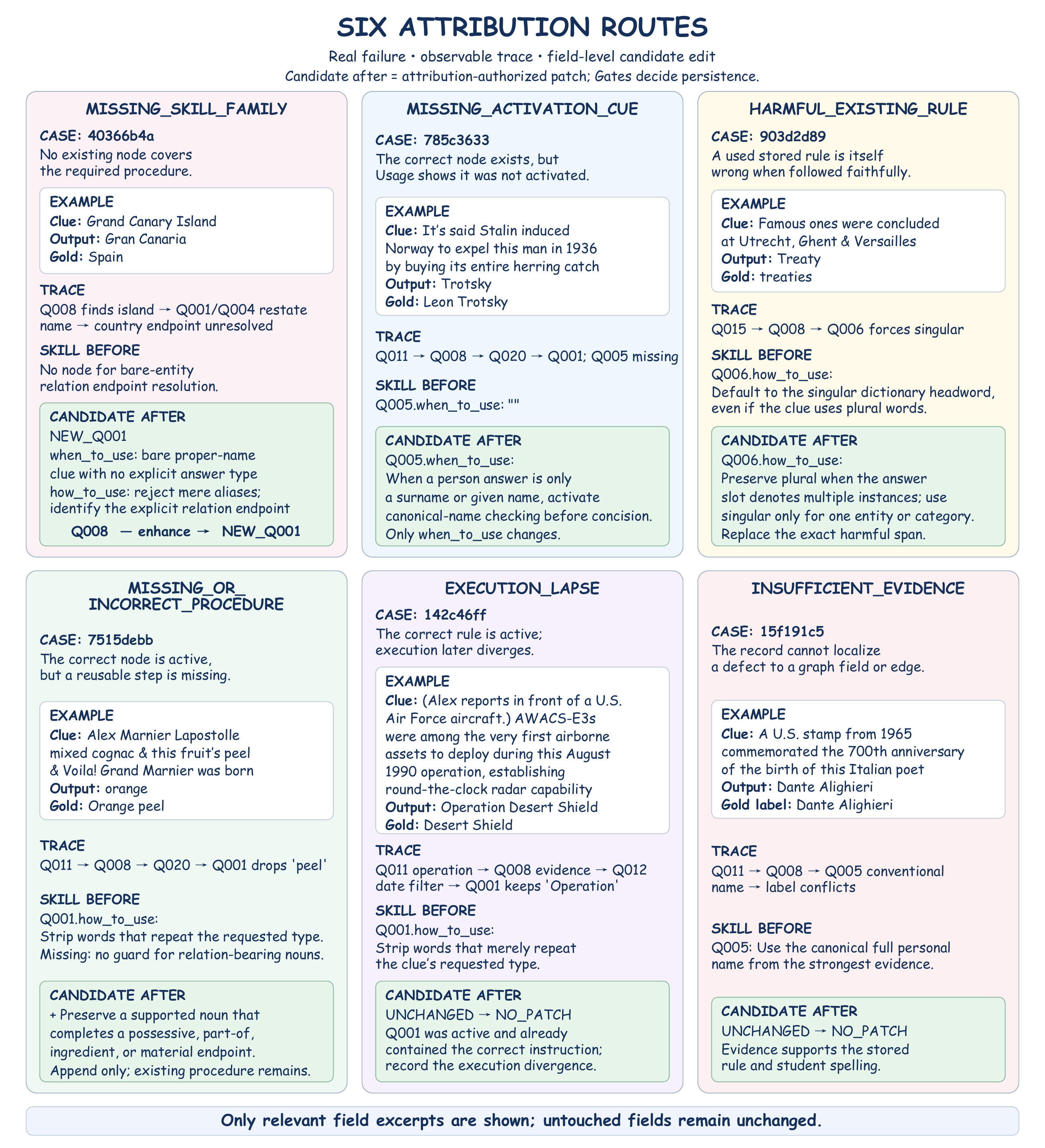}
\setlength{\abovecaptionskip}{3pt}
\setlength{\belowcaptionskip}{0pt}
\caption{Six attribution routes and examples of field-level edits. Each card gives a SearchQA failure's question, prediction, gold answer, checked execution record, and the relevant fields before and after editing. \emph{Candidate after} awaits Gate validation, and only changed fields are shown. \texttt{EXECUTION\_LAPSE} and \texttt{INSUFFICIENT\_EVIDENCE} retain the original graph.}
\label{fig:six-attribution-routes}
\end{figure*}

Figure~\ref{fig:six-attribution-routes} illustrates the six mutually exclusive attribution outcomes. The first four correspond to node-field or relation edits, while \texttt{EXECUTION\_LAPSE} and \texttt{INSUFFICIENT\_EVIDENCE} return \texttt{NO\_PATCH}.

These edits are validated as repair groups. Before--after results on source and affected examples determine whether each group is retained or rolled back.

\subsection{Gate Scope Expansion and Rollback Mechanism}
\label{app:local-gate-details}

After identifying the edit targets, the system uses usage records and graph relations to select examples for retesting, and retains or rolls back dependent operations together.

The Local Gate for repair group $J_k$ retests four sets of examples from the current update pool:
\begin{itemize}
\item $\mathrm{DirectSources}_k$: failed examples cited as source evidence by an operation in the group;
\item $\mathrm{PriorUsers}_k$: examples whose valid pre-edit \texttt{usage} includes an edited object, matching the node ID for a node-field edit or the edge ID and both endpoint IDs for an existing-edge edit;
\item $\mathrm{UnknownUsage}_k$: examples with missing or invalid pre-edit \texttt{usage} records; and
\item $\mathrm{GroupSiblings}_k$: other update examples in the same update group as any example in the preceding three sets.
\end{itemize}
The retest set is
\[
\mathrm{Retest}_k=
\mathrm{DirectSources}_k
\cup\mathrm{PriorUsers}_k
\cup\mathrm{GroupSiblings}_k
\cup\mathrm{UnknownUsage}_k.
\]

When adding an edge between two existing nodes, $\mathrm{PriorUsers}_k$ includes examples whose pre-edit usage records contain either the source or the target node.

For a group that only adds a specialist node and its connecting edge, the new node has no prior users, so $\mathrm{PriorUsers}_k$ is empty. Local retesting includes source examples, usage-unknown examples, and their group siblings. The Big Gate checks the parent's other users as part of the complete update pool.

Comparing candidate and pre-edit reference results example by example gives three counts: $n_{\mathrm{fixed}}^k$ counts incorrect-to-correct answers, $n_{\mathrm{broken}}^k$ counts correct-to-incorrect answers, and $n_{\mathrm{unresolved}}^k$ counts unresolved direct source failures. Acceptance requires
\[
n_{\mathrm{fixed}}^k
>
n_{\mathrm{broken}}^k
+
n_{\mathrm{unresolved}}^k.
\]
Validation applies to the complete $J_k$. For example, a new node and its connecting edge are checked together rather than selected separately. Each corrected error contributes $+1$, each broken success contributes $-1$, and each unresolved source failure contributes another $-1$. A positive score retains the group; otherwise it proceeds to refinement or rollback. This condition accounts for both new errors and unresolved source failures, allowing some regressions when the repair gains are sufficient.

The Local Gate decides from these counts, and the teacher then explains the results for audit. If the first attempt fails, the graph structure is valid, and retesting reveals cases where the repair is ineffective, the group may be refined once within the original target scope and retested. The same acceptance condition applies to the second attempt. A second failure rolls back the whole group.

\paragraph{Paired before--after comparison.}
Each retested example is compared with its pre-edit reference result. Both sides use the same example identity and order, model route, reasoning effort, output limit, and response settings, changing only the rendered graph. The Local Gate compares affected examples, while the Big Gate compares the complete update pool.

\paragraph{Whole-graph rollback.}
Repair groups that pass the merge checks are applied to the old graph and evaluated on the complete update pool. If corrected cases do not outnumber broken cases, the merged candidate is discarded. The old graph, its prompt, and the epoch-start reference results are restored for the next epoch.

\paragraph{Coverage and outside regressions.}
Local-Gate coverage is the fraction of distinct update examples retested in each round. The Big-Gate outside regression rate is the fraction of assessed examples outside Local-Gate coverage whose answers change from correct to incorrect. Case counts and coverage are averaged over rounds with completed candidate evaluations. The regression rate divides the total regressions across those rounds by the total number of assessed examples outside Local-Gate coverage.

\begin{table}[H]
\centering
\footnotesize
\renewcommand{\arraystretch}{1.12}
\begin{tabular*}{\textwidth}{@{\extracolsep{\fill}}lrrr}
\toprule
Dataset & \makecell[r]{Outside cases\\(mean per round)} & \makecell[r]{Local Gate\\coverage} & \makecell[r]{Big Gate outside\\regression rate} \\
\midrule
SearchQA & 352.5 & 41.25\% & 3.40\% \\
LiveMath & 253.5 & 27.78\% & 9.86\% \\
DocVQA & 562 & 6.33\% & 3.02\% \\
\bottomrule
\end{tabular*}
\caption{Local-Gate coverage and Big-Gate outside regression rates for GPT-5.6-sol, with Big Gate evaluated on subsets.}
\label{tab:gate-coverage-supplement}
\end{table}

Mean Local-Gate coverage ranges from 6.33\% to 41.25\%, concentrating on source examples and those linked by usage records. After merging, 3.02\%--9.86\% of the assessed examples outside that scope change from correct to incorrect. The Big Gate detects these regressions and includes them in the merged candidate's acceptance decision.

LiveMath has the highest outside regression rate, at 9.86\%. Although DocVQA has the lowest local coverage, at 6.33\%, it also has the lowest outside regression rate, at 3.02\%. Greater coverage does not directly imply fewer regressions: local retesting uses records of which rules were used before editing, while the updated graph may lead the model to select different rules, and merged repairs may introduce new interactions. The merged graph therefore still needs to be evaluated after group-wise checks.

\subsection{Validation Checks and Outcomes}
\label{app:validation-levels}

Validation checks record integrity, the validity of edit locations, local behavioral improvement, and the overall gain of the merged graph. Examples with missing or invalid records are included in retesting as usage-unknown, as summarized in Table~\ref{tab:validation-levels}.

\begin{table}[ht]
\centering
\footnotesize
\renewcommand{\arraystretch}{1.20}
\setlength{\tabcolsep}{4pt}
\renewcommand{\tabularxcolumn}[1]{m{#1}}
\begin{tabularx}{0.94\textwidth}{>{\raggedright\arraybackslash}m{0.16\textwidth}>{\raggedright\arraybackslash}X>{\raggedright\arraybackslash}X}
\toprule
\rowcolor{tableheadergray}
\multicolumn{1}{c}{\textbf{Check type}} & \multicolumn{1}{c}{\textbf{Check}} & \multicolumn{1}{c}{\textbf{Outcome}} \\
\midrule
Record integrity & IDs in \texttt{trace} and \texttt{usage} agree, exist in the current graph, and have valid endpoints. & The record can determine the affected scope by graph object; otherwise usage is marked unknown and the retest scope is expanded. \\
\lightrowrule
Routing validity & The earliest observable error, successful contrasts, graph coverage, activation, stored semantics, and execution are checked in order. & The node fields and relations to edit are identified, or \texttt{NO\_PATCH} is returned when localization fails. \\
\lightrowrule
Local behavioral validity & The complete atomic group is rerun on direct sources, prior users, group siblings, and usage-unknown cases. & The Local Gate decides whether to retain or revert the group. \\
\lightrowrule
Global behavioral validity & All locally retained groups are merged and rerun on the complete update pool. & The merged graph is committed if corrected cases outnumber broken cases. \\
\bottomrule
\end{tabularx}
\caption{Validation checks and outcomes.}
\label{tab:validation-levels}
\end{table}

\section{Prompt Templates}
\label{app:prompts}

\subsection{From Initial Skills to Graphs: Conversion and Execution}
\label{app:searchqa-graph-workflow}
\label{app:skill-to-graph}

\paragraph{Initialization and semantic conversion.}
\GraphSkillAA{} first initializes a task skill document $S$, which an LLM converts into the initial graph $G^{(0)}$. Here, the best SkillOpt skill supplies the initialization, and Flat skill and the graph variants use the same source document.

\paragraph{From rules to nodes.}
The LLM organizes nodes by rule meaning: rules sharing conditions and purpose can be grouped, while independent checks form separate nodes. \texttt{when\_to\_use} records applicability, \texttt{how\_to\_use} records the procedure, and \texttt{avoid} records inapplicable situations or actions to avoid. Conditions closely tied to an action may remain in \texttt{how\_to\_use}; fields not separately specified in the source remain empty.

\paragraph{From relationships between rules to edges.}
The LLM also interprets the source skills to determine which nodes to connect and the direction and type of each edge. If B requires A to be completed first, a \texttt{prereq} edge connects source A to target B. If A helps B execute better, an \texttt{enhance} edge expresses that assistance. If two rules are used together without an ordering requirement, \texttt{co\_occur} expresses their association. Nodes retain the organized skill content, while edges make the coordination identified by the LLM explicit.

\paragraph{Box D.1: From table-lookup rules to an initial subgraph.}
The DocVQA table-lookup example below follows the complete conversion, with source rules and field text abbreviated.

\paragraph{1. Source skill.}
The source document contains the following requirements:
\begin{itemize}
\item In a table, find the row named in the question, read the requested column, and return only that cell's value.
\item When nearby strings look similar, choose the one whose labels or layout best match the question.
\item Copy names, numbers, and dates exactly from the document whenever possible.
\item Preserve spelling and punctuation without substituting similar letters or changing quotes, spacing, or parentheses.
\item Prefer the smallest exact text span that answers the question.
\end{itemize}

\paragraph{2. Split rules and assign fields.}
The LLM separates cell lookup, candidate verification, exact copying, and answer-span selection into four nodes. The third and fourth requirements both govern copying and are grouped into D003. Applicability goes into \texttt{when\_to\_use}, actions into \texttt{how\_to\_use}, and prohibited alterations into \texttt{avoid}:

\begin{promptbox}[Box D.1: field mappings]
[D005] Table row-and-column lookup
when_to_use: For tables.
how_to_use: Find the named row, then the requested column or header.
            Return only that cell's value.
avoid: []

[D002] Candidate verification
when_to_use: Several nearby strings look similar.
how_to_use: Choose the one whose labels or layout best match the question.
avoid: []

[D003] Exact copying
when_to_use: ""
how_to_use: Copy names, numbers and dates exactly.
            Preserve spelling and punctuation.
avoid: Do not substitute similar letters or alter quotes,
       spacing or parentheses visible in the document.

[D009] Answer span selection
when_to_use: ""
how_to_use: Prefer the smallest exact span that answers the question.
avoid: []
\end{promptbox}

Here, D005's table setting is an applicability condition. D003's exact copying is an action, while its prohibition on altering characters constrains that action. Fields remain empty when no separate condition or prohibition is specified.

\paragraph{3. Connect related rules.}
The LLM then determines how the rules work together, rather than simply linking nodes in source-document order:
\begin{itemize}
\item Table lookup must supply a candidate cell before its labels and layout can be checked, giving \texttt{E005: D005 -[prereq]-> D002}.
\item Checking similar candidates helps select the correct answer, giving \texttt{E002: D002 -[enhance]-> D009}.
\item Preserving source characters helps produce an exact answer span, giving \texttt{E003: D003 -[enhance]-> D009}.
\end{itemize}
The resulting subgraph contains four nodes and three edges. D002 and D003 assist D009 by selecting the right content and preserving its characters, respectively. They need not form a sequential chain.

\paragraph{4. Form the student input.}
The nodes and edges are saved together and rendered with the fixed task protocol into the student prompt. The question and document image provide the task input. For illustration, suppose a table lists unit prices of \texttt{120.00} for A01 and \texttt{210.00} for A10, and the question asks for A01's unit price. D005 guides row-and-column lookup, D002 helps distinguish A01 from A10, D003 preserves the exact text \texttt{120.00}, and D009 directs the student to return only that value. The example shows how the saved rules and their relations jointly guide an answer.

\paragraph{Storage, execution, and updating.}
The program checks fields, referenced IDs, and acyclicity of the \texttt{prereq} subgraph, then saves the nodes and edges. Runs start from the same initial graph. The renderer combines the current graph in fixed order with the task protocol and output contract. Determinism refers to rendering a given graph; the LLM makes the semantic judgments in initial construction. The student interprets and uses the relations in the prompt, while the program does not invoke rules one by one along edges. Subsequent teacher updates use execution feedback to revise nodes and \texttt{prereq}/\texttt{enhance} relations, with commitment following two-level Gate acceptance. The initial \texttt{co\_occur} relations are retained.

\begin{figure}[H]
\centering
\includegraphics[width=0.96\textwidth]{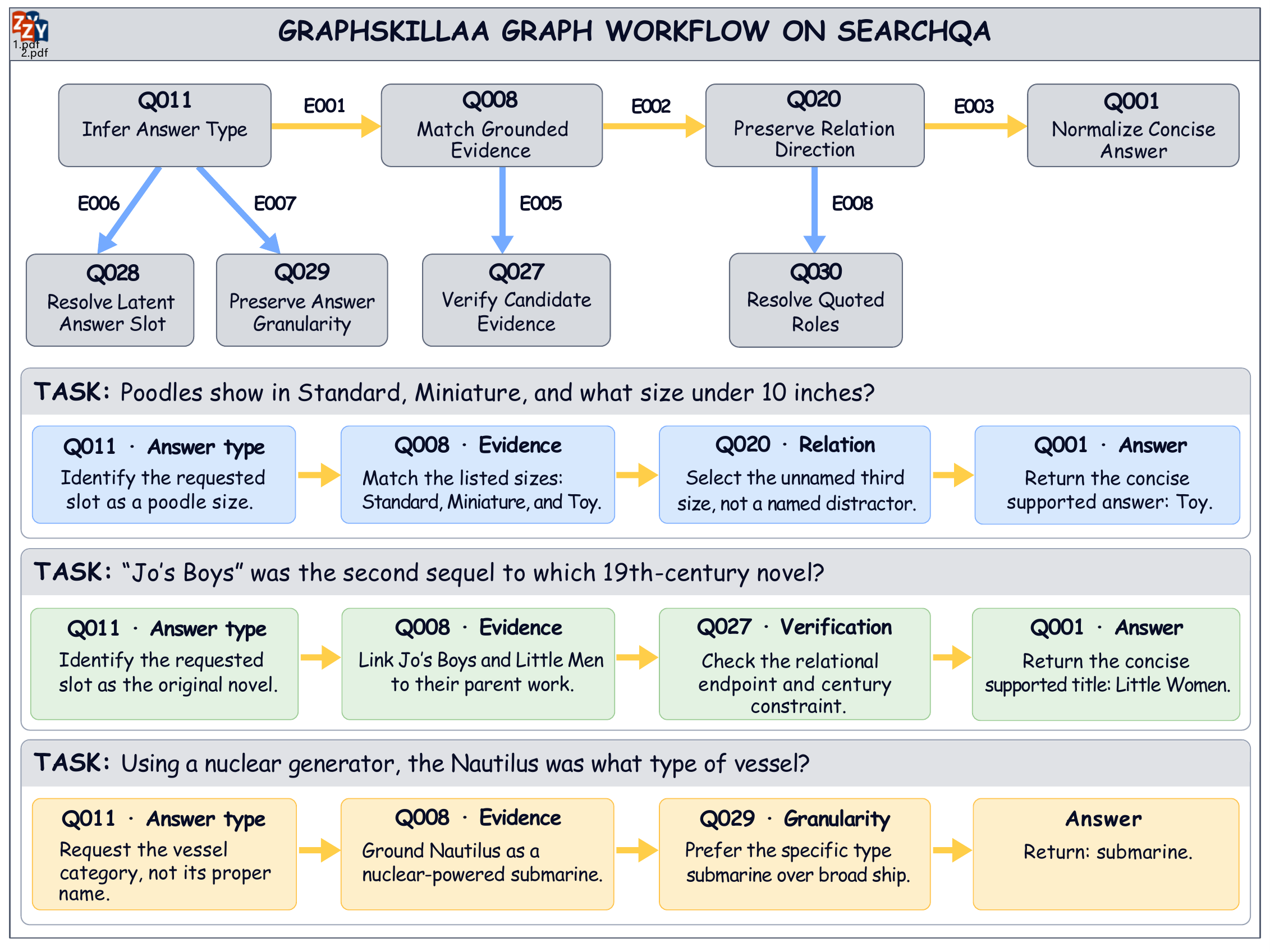}
\setlength{\abovecaptionskip}{3pt}
\setlength{\belowcaptionskip}{0pt}
\caption{GraphSkillAA graph workflow on SearchQA. The upper panel shows the core prerequisite chain and specialist enhancement edges; the lower panels summarize decisions for three questions, with arrows indicating presentation order.}
\label{fig:searchqa-graph-workflow}
\end{figure}

The upper panel of Figure~\ref{fig:searchqa-graph-workflow} shows eight nodes and their relations from the optimized SearchQA graph, while the lower panel shows key decisions for three questions. Node IDs link rules used during execution to edit locations in the graph.

In the third example, Q029's granularity check is linked to Q011 through E007. The lower-panel decision summaries follow presentation order, with complete execution records in the student's \texttt{trace} and \texttt{usage}.

\clearpage

\subsection{Abbreviated SearchQA SkillGraph Example}
\label{app:skillgraph-example}

The following excerpt shows the SearchQA SkillGraph received by the student. Both example nodes contain three fields, edge rationales are abbreviated, and ellipses mark the remaining nodes and relations.

\begin{promptbox}[Abbreviated SearchQA SkillGraph example]
Name: SearchQA SkillGraph
Task description: Answer a short open-domain question from the supplied
retrieved passages and return the shortest supported answer.

General principles (fixed task protocol):
- Answer only from the supplied context.
- Infer the requested answer type before extracting a candidate.
- Return reasoning_trace, graph_usage, and answer in the required format.
- Only the graph may evolve; never memorize a case ID or gold answer.

Node set V:
[Q001] Concise Answer Normalization 1
When to use: A supported candidate must be converted into the final
short-answer surface form.
How to use: Return the shortest unambiguous supported answer; remove only
generic descriptors that are not needed for identification.
Avoid: Do not shorten a full official name when it is explicitly requested
or when the shorter form would be ambiguous.

...

[Q027] Candidate Evidence Verification
When to use: Candidate evidence is ambiguous, partial, conflicting, or
supports a related entity of the wrong type.
How to use: Verify the requested type, endpoint, and every explicit
constraint before accepting the candidate.
Avoid: Do not replace a candidate unless another candidate satisfies the
complete constraint set with direct evidence.

Edge set E:
[E004] Q005 -[enhance; strength=strong]-> Q026
  Rationale: Keep the evidence-backed exception attached to Q005 without
  changing the base-node semantics globally.
[E005] Q008 -[enhance; strength=strong]-> Q027
  Rationale: Confine candidate verification to observable ambiguity instead
  of changing Q008 globally.
...
\end{promptbox}

\clearpage
\subsection{Student Execution and Trace/Usage Specification}
\label{app:prompt-student}

In addition to its answer, the student produces two records: \texttt{trace} describes execution, while \texttt{usage} lists the node and edge IDs the model reports using to identify examples that edits may affect.

\begin{promptbox}[Student execution contract]
You are an expert question answering agent.
Answer only from the supplied context.

Before the final answer, write exactly one plain-text graph execution
record inside <reasoning_trace>...</reasoning_trace>. It must expose the
semantic rule path, not merely summarize the task:

- For every applied node, cite its exact graph ID, state the evidence or
  intermediate state entering that node, what the node did, and the new
  intermediate result.
- For every followed edge, cite its exact edge ID, name its endpoints,
  explain why the dependency applied, and state what it enabled.
- State the requested answer type, decisive evidence, relation or
  constraint, rejected endpoint when relevant, and final surface form.
- Do not cite an object that was not actually applied.

Output exactly these blocks in order:
<reasoning_trace>...</reasoning_trace>
<graph_usage>{"used_nodes":["..."],"used_edges":["..."]}</graph_usage>
<answer>...</answer>

graph_usage must be the exact deduplicated set of IDs explicitly cited
as applied in reasoning_trace. Use only IDs visible in the current graph.
Only the graph may evolve; do not memorize case IDs or gold answers.
\end{promptbox}

\clearpage
\subsection{Abductive Attribution Prompt}
\label{app:prompt-attribution}

This prompt first assigns a verified failure to one root-cause category and then lists all fields or relations to be changed, including necessary dependent operations. The field \texttt{retrieval\_revision\_proposals} denotes edits to \texttt{when\_to\_use} for existing nodes.

\begin{promptbox}[Case analyzer: root cause and edit routing]
You are the GraphOpt Case Analyzer. Read the frozen SkillGraph G_t and
one complete case record, then perform semantic node- and edge-level
attribution. Return exactly one parseable JSON object.

Evidence:
- Inspect the original task, response, evaluator outcome, semantic trace,
  graph_usage, and successful siblings together.
- A validated student trace is an execution record bound to exact graph IDs
  and explicit input--operation--output transitions.
- VERIFIED usage requires exact agreement between cited IDs, the sidecar,
  and valid graph endpoints.

First locate the earliest observable decision after which the answer is
unrecoverable. Then choose exactly one root-cause route:

SUCCESS -> no edit.
MISSING_ACTIVATION_CUE -> clarify only target.when_to_use.
MISSING_OR_INCORRECT_PROCEDURE -> PATCH only target.how_to_use or
  target.avoid, or add the necessary prereq/enhance relation.
HARMFUL_EXISTING_RULE -> replace only the exact harmful span, or atomically
  replace the incorrect edge.
MISSING_SKILL_FAMILY -> add one complete specialist node and an enhance
  edge from a trace-used correct parent in the same atomic group.
EXECUTION_LAPSE -> record the divergence and emit no graph proposal.
INSUFFICIENT_EVIDENCE -> emit no graph proposal.

Every proposed edit must implement the smallest reusable contrast with
successful cases. Never copy a gold answer, case ID, document name, or
hindsight-only condition into a skill. A missed node does not by itself
justify broadening its trigger.

List every target node, field, and relation in the proposals, including
dependent operations such as a new node and its connecting edge. Joint
synthesis will complete the edits within these specified targets.

Required output fields:
case_id, success, failure_type, badcase_summary,
existing_graph_usage, node_revision_proposals,
retrieval_revision_proposals, edge_correction_proposals,
new_node_proposals, new_edge_proposals.
\end{promptbox}

\clearpage
\subsection{Epoch-Level Meta-Auditor Prompt}
\label{app:prompt-meta}

The epoch-level meta-auditor summarizes which edits worked and which failed, providing guidance for later epochs. The two Gates still decide whether to commit the graph.

\begin{promptbox}[Epoch meta auditor]
You are the GraphOpt Epoch Meta Auditor.

Resolve every exact evidence reference through evidence_catalog verbatim
before auditing. The input contains previous_meta, previous and current
graph summaries, longitudinal improved/regressed/persistent-fail/stable-
success pairs, and complete Local- and Big-Gate records.

Use only this input. Read the complete Gate measurements, edit attribution,
original tasks, responses, trajectories, references, and before/after
states together. Independently verify that each lesson is supported by
the original evidence; do not rely on a decision summary when complete
evidence contradicts it.

Preserve every non-contradicted, evidence-grounded lesson in previous_meta,
then add or revise lessons supported by the current epoch. Cover effective
edits, regressions, likely missing skills, and changes to avoid. The
Permanent Protocol is frozen; statistics remain in evolution_cache;
successful cases cannot propose edits; this channel cannot propose
co_occur.

Return exactly:
{"schema_version":"graphopt-meta-v1",
 "bullets":["specific evidence-grounded point", "..."]}

Return 5--12 unique non-empty strings, each at most 500 characters. If
evidence is sparse, state uncertainty and the conservative action instead
of inventing a conclusion.
\end{promptbox}

\clearpage
\subsection{Patch Synthesis and Gate Audit Prompts}
\label{app:prompt-patch-gate}

The first template adds a conditional operation to an existing node while preserving its original field content. Other attribution routes use their corresponding field templates. The second template explains the already determined Local-Gate result for audit.

\begin{promptbox}[Targeted node-patch synthesizer]
Synthesize the complete epoch evidence for one existing node into one
precise conditional lesson. The program appends it to the Original Node;
you are not allowed to rewrite the node.

Return exactly {"addition":"..."} or {"addition":null}.

- Inspect original_task, student_output, evaluation, training reference,
  reasoning evidence, graph_usage, proposed changes, and Successful Uses.
- Treat failures and successes as a paired counterfactual set. Identify
  the smallest observable state separating them.
- State observable condition -> next action/state update, with a narrow
  exception when needed. Preserve every byte of the old node.
- Do not create unconditional always/never rules from failure-only
  patterns. Keep the addition under 80 words.
- Return null if the proposal is already covered, targets another node,
  conflicts with a Successful Use, or lacks an observable condition.
\end{promptbox}

\begin{promptbox}[Local-Gate explanation auditor]
Explain one measured local joint graph edit from the supplied paired case
summaries and expanded decisive/protected traces. The deterministic Gate
decision is already final; do not change it.

Use only the supplied questions, reasoning traces, graph_usage, answers,
and hard outcomes. Do not claim that one edit was independently effective
when node/edge edits were intervened on jointly. Distinguish observations
from hypotheses.

Return strict JSON with joint_summary and one entry per edit containing
edit_index, effective_mechanism, ineffective_mechanism,
observable_boundary, and evidence_case_ids.
\end{promptbox}

\clearpage

\section{Experimental Protocol}
\label{app:protocol}

This appendix describes the data splits, model settings, and scoring methods for Tables~\ref{tab:table-a}--\ref{tab:table-d} and Figure~\ref{fig:epoch-trajectories}.

\subsection{Dataset Splits and Group Construction}
\label{app:benchmark-splits}

The data are divided into an update pool for modifying the skill graph and a test set for evaluation. Table~\ref{tab:split-counts} lists both protocols.

\begin{table}[ht]
\centering
\renewcommand{\arraystretch}{0.96}
\setlength{\tabcolsep}{4.5mm}
\footnotesize
\resizebox{0.60737\textwidth}{!}{%
\begin{tabular}{lrrr}
\toprule
\rowcolor{tableheadergray}
\textbf{Protocol / benchmark} & \textbf{Train} & \textbf{Validation} & \textbf{Test} \\
\midrule
Main / SearchQA & 800 & 0 & 200 \\
Main / LiveMath & 468 & 0 & 117 \\
Main / DocVQA & 800 & 0 & 200 \\
SkillOpt / SearchQA & 400 & 200 & 1,400 \\
SkillOpt / LiveMath & 35 & 18 & 124 \\
SkillOpt / DocVQA & 107 & 53 & 374 \\
\bottomrule
\end{tabular}%
}
\caption{All benchmark split counts. Main-protocol ``Train'' is the complete update pool.}
\label{tab:split-counts}
\end{table}

The main protocol forms fixed groups of four update examples. Any remainder is kept in the last batch so that each epoch covers all training examples. Grouping uses update-question text and task type: SearchQA uses BGE cosine similarity, DocVQA additionally requires compatible question types, and LiveMath first separates fine-grained mathematical categories and then groups by TF--IDF similarity. Test inputs, answers, and run results never enter grouping. In the SkillOpt protocol, train and validation form the update pool, while the test set is used for evaluation. Every DocVQA setting uses the original image and question.

\subsection{Model Access, Inference, and Optimization Settings}
\label{app:model-settings}

\paragraph{Models.}
The main protocol uses GPT-5.6-sol, Gemini-3.5-Flash, and Qwen3.8-Flash, with each model serving as both student and teacher (the self-teacher setting). The SkillOpt-protocol comparison uses GPT-5.5, GPT-5.4-mini, and Qwen3.6-35B-A3B under the same arrangement. Final testing calls only the student. Teacher-transfer experiments in Table~\ref{tab:table-c} hold the student fixed and use GPT-5.4-mini and GPT-5.6-sol as alternative teachers. The release will identify the model version used in each run.

\paragraph{Inference.}
All models are called through APIs with \texttt{medium} reasoning effort. Student answers use a 16,384-token output limit. Teacher output limits vary by stage: 768--4,096 tokens for attribution, audit, and refinement, and up to 8,192 tokens for synthesis and merging. Qwen thinking is enabled for SearchQA and DocVQA and disabled for the LiveMath fast protocol. Model outputs are subject to the provider's sampling mechanism. Each run records the model version, access method, decoding settings, timestamps, skill graph, and raw outputs.

\paragraph{Optimization.}
The three independent runs use seeds 42, 43, and 44. \GraphSkillAA{} and its update-mechanism ablations in Tables~\ref{tab:table-a}--\ref{tab:table-c} run for three epochs over the complete update pool. The other baselines in Table~\ref{tab:table-b} retain their own optimization procedures. Retrieval ablations use the optimized graph and change only the content presented to the student during evaluation. Gate scope and acceptance follow Section~\ref{sec:method} and Appendix~\ref{app:local-gate-details}.

\paragraph{Optimization and deployment cost.}
Table~\ref{tab:optimization-cost} reports API token usage for one complete optimization run of each method under the GPT-5.6-sol main protocol. For \GraphSkillAA{}, this covers student execution, teacher attribution, joint patch synthesis, patch refinement, and Local Gate/Big Gate validation, starting from the given initial skill and excluding upstream SkillOpt training. Accuracy uses the GPT-5.6-sol three-test means in Table~\ref{tab:table-a}; SkillOpt corresponds to Flat skill, which directly uses SkillOpt's released best skills.

\begin{table}[H]
\centering
\small
\renewcommand{\arraystretch}{1.14}
\setlength{\tabcolsep}{3pt}
\begin{tabularx}{\textwidth}{@{}ll*{5}{>{\raggedleft\arraybackslash}X}@{}}
\toprule
\rowcolor{tableheadergray}
\textbf{Dataset} & \textbf{Method}
& \makecell[r]{\textbf{Actual}\\\textbf{Tokens (M)}}
& \makecell[r]{\textbf{Token}\\\textbf{Ratio}}
& \makecell[r]{\textbf{Accuracy}\\\textbf{(\%)}}
& \makecell[r]{\textbf{Gain}\\\textbf{(pp)}}
& \makecell[r]{\textbf{Extra Tokens}\\\textbf{/ pp (M)}} \\
\midrule
SearchQA & SkillOpt & 25.09 & $1.00\times$ & 72.5 & --- & --- \\
 & GraphSkillAA & 53.18 & $2.12\times$ & \textbf{81.5} & \textbf{+9.0} & 3.121 \\
\lightrowrule
LiveMath & SkillOpt & 0.50 & $1.00\times$ & 42.2 & --- & --- \\
 & GraphSkillAA & 1.78 & $3.56\times$ & \textbf{66.7} & \textbf{+24.5} & 0.052 \\
\lightrowrule
DocVQA & SkillOpt & 1.13 & $1.00\times$ & 87.7 & --- & --- \\
 & GraphSkillAA & 4.37 & $3.87\times$ & \textbf{91.2} & \textbf{+3.5} & 0.926 \\
\bottomrule
\end{tabularx}
\caption{Actual optimization token usage per complete run and accuracy with GPT-5.6-sol. M denotes millions of tokens.}
\label{tab:optimization-cost}
\end{table}

Token Ratio gives optimization tokens relative to SkillOpt, and Gain is the accuracy improvement in percentage points. Extra Tokens / pp is the difference between the two workflows' optimization tokens divided by their accuracy difference: (GraphSkillAA tokens $-$ SkillOpt tokens) / (GraphSkillAA accuracy $-$ Flat skill accuracy), in millions of tokens per percentage point.

Measured total deployment costs are similar. Most deployment steps are shared: both methods use the same student model, inference settings, and test inputs, load fixed skills, and perform one student inference round per question. GraphSkillAA's teacher attribution, patch generation, and both Gates run during optimization and add no deployment calls.

Both methods use model APIs, so elapsed time depends on server load, network conditions, and concurrency. We measure computational cost using actual tokens consumed by the full optimization run. Extra Tokens / pp relates the difference in the two workflows' token use to their accuracy difference.

\GraphSkillAA{} further optimizes SkillOpt's released best skills. With GPT-5.6-sol, accuracy improves by $9.0$, $24.5$, and $3.5\pp$ on SearchQA, LiveMath, and DocVQA, respectively, over directly using these text skills. Thus, finer-grained attribution, targeted edits, and two-level validation can further raise attainable skill-optimization performance without updating model parameters. We consider this additional computation worthwhile for performance-first applications.

\subsection{Evaluation Protocol}
\label{app:artifact-provenance}

SearchQA and LiveMath use exact-match accuracy. DocVQA counts an answer as correct when ANLS$=1$ and reports the proportion of correct answers. The Gates use the same scoring rule to count incorrect-to-correct ($0\!\to\!1$) and correct-to-incorrect ($1\!\to\!0$) changes. Graph-optimization settings produce and evaluate one final committed graph for each seed, 42, 43, and 44. Frozen settings keep the skill fixed and repeat evaluation only. The three evaluations share the scorer, model access method, reasoning effort, and output budget. We report their mean and half-range, $(\max-\min)/2$. Test results do not enter attribution, updating, or graph selection. The epoch-level test results in Figure~\ref{fig:epoch-trajectories} describe performance after updating.

\subsection{Result Sources and Experimental Settings}
\label{app:comparison-ledger}

Table~\ref{tab:comparison-ledger} summarizes the optimization targets, teacher assignments, and evaluation settings for each experiment. Table~\ref{tab:table-b} follows the direct-chat setting of SkillOpt Table~1~\citep{skillopt2026}: methods use the same released examples, test IDs, original inputs, inference budget, and scorer while retaining their own optimization procedures. \GraphSkillAA{} updates on train $\cup$ validation and is scored on the test set.

\begin{table}[ht]
\centering
\footnotesize
\renewcommand{\arraystretch}{1.16}
\setlength{\tabcolsep}{3.2pt}
\renewcommand{\tabularxcolumn}[1]{m{#1}}
\begin{tabularx}{0.97\textwidth}{>{\raggedright\arraybackslash}m{0.18\textwidth}>{\raggedright\arraybackslash}m{0.21\textwidth}>{\raggedright\arraybackslash}m{0.18\textwidth}>{\raggedright\arraybackslash}X}
\toprule
\rowcolor{tableheadergray}
\multicolumn{1}{c}{\textbf{Block}} & \multicolumn{1}{c}{\textbf{Optimized state}} & \multicolumn{1}{c}{\textbf{Teacher assignment}} & \multicolumn{1}{c}{\textbf{Evaluation setting}} \\
\midrule
Table~\ref{tab:table-a}, frozen stages & None; the flat skill or G0 is fixed & No teacher call & Repeated evaluation of fixed skills on the same test inputs. \\
\lightrowrule
Table~\ref{tab:table-a}, GraphSkillAA & Three-epoch graph optimization & Student is its own teacher & Seeds 42/43/44 produce three committed graphs evaluated on the test set. \\
\lightrowrule
Table~\ref{tab:table-b}, other baselines & Method-specific states from the original paper & Original reported settings & Results reported under the SkillOpt direct-chat setting. \\
\lightrowrule
Table~\ref{tab:table-b}, GPT-5.5 SkillOpt & Released best skills, frozen & No teacher calls & Mean of three local evaluations using the original test examples and scoring method. \\
\lightrowrule
Table~\ref{tab:table-b}, GraphSkillAA & Three-epoch graph optimization on train $\cup$ validation & Student is its own teacher & GraphSkillAA evaluation under the same direct-chat setting and on the same test set. \\
\lightrowrule
Table~\ref{tab:table-c}, training interventions & Three-epoch graph optimization & Self-teacher unless the row names a teacher & Each configuration is optimized separately, with final committed graphs evaluated on the same test set. \\
\lightrowrule
Table~\ref{tab:table-c}, retrieval interventions & Final graph from three-epoch self-teacher optimization; intervention only at evaluation & Student is its own teacher during optimization; no teacher call at evaluation & The source graph and test examples remain fixed; only graph content presented to the student changes. \\
\lightrowrule
Table~\ref{tab:table-c}, teacher intervention & Three-epoch graph optimization & GPT-5.4-mini or GPT-5.6-sol as named & Student, data, edit rules, epoch count, and test settings remain fixed. \\
\bottomrule
\end{tabularx}
\caption{Settings of the reported experimental blocks.}
\label{tab:comparison-ledger}
\end{table}

\section{Ablation Protocols and Retrieval Definitions}
\label{app:ablations}

Table~\ref{tab:table-c} examines graph structure, skill retrieval, updating, and teacher models. Evaluation examples, original inputs, scorers, and student inference settings remain the same across variants.

\subsection{Shared Settings and Graph-Structure Ablation}
\label{app:graph-ablation}

\paragraph{Flat skill.}
Flat skill uses the original SkillOpt prompt without graph conversion or optimization. Its results in Table~\ref{tab:table-c} come from an independent evaluation using the same split, inputs, scorer, and inference settings as Table~\ref{tab:table-a}, again over three runs.

\paragraph{Nodes only.}
Nodes only removes every edge from the corresponding optimized graph while keeping all nodes and their fields unchanged. Its difference from GraphSkillAA (Full) measures the contribution of edges.

\paragraph{Information provided by graph relations.}
Nodes only retains the complete structured skill content, so its comparison with GraphSkillAA (Full) tests relationship information between skills. For GPT-5.6-sol, removing edges lowers accuracy by $4.0$, $4.6$, and $1.2\pp$ on SearchQA, LiveMath, and DocVQA, respectively. Hiding only edge explanations while retaining endpoints, types, directions, strengths, order, and all node fields (Appendix~\ref{app:retrieval}) lowers accuracy by $3.0$, $3.2$, and $2.2\pp$, respectively. These two evaluation-time interventions show that explicit relations and their explanations help the model combine skills when node content is held fixed.

\subsection{Retrieval-Mechanism Ablations}
\label{app:retrieval}

Retrieval ablations use each student's final graph from the self-teacher setting and change only the content presented during testing. Evaluation examples, generation settings, and original inputs remain fixed, and DocVQA always receives the complete image and question.

\paragraph{No activation boundaries.}
The renderer omits only \texttt{when\_to\_use} and \texttt{avoid}. Node IDs, titles, \texttt{how\_to\_use}, and all edge fields remain unchanged and in the same order.

\paragraph{No edge rationale.}
The renderer omits only each edge's \texttt{rationale}. Endpoints, type, direction, strength, order, and all node fields remain unchanged.

\paragraph{S--Q retrieval.}
This variant follows the embedding-retrieval core of SkillRL~\citep{skillrl2026}. The fixed encoder \texttt{Qwen/Qwen3-Embedding-0.6B} embeds the current question $q$ and node text $s_j$, formed by concatenating title, \texttt{how\_to\_use}, and \texttt{when\_to\_use}. Normalized inner product ranks nodes:
\[
r_j(q)=f(q)^\top f(s_j),\qquad
R_{\mathrm{SQ}}(q)=\operatorname{TopK}_{j}r_j(q),\quad K=6.
\]
The \texttt{avoid} field does not affect ranking but is rendered for selected nodes. Nodes outside the top six are omitted, and an edge is retained only when both endpoints are selected.

\paragraph{Nearest-Q reuse.}
This variant uses the same test set as GraphSkillAA (Full). For a test question $q$, it selects the most similar question from the update pool:
\[
i^*=\arg\max_{i\in\{1,\ldots,|\mathcal D_{\mathrm{update}}|\}}
\operatorname{sim}_d(q,q_i^{(u)}).
\]
Here, $\operatorname{sim}_d$ follows the similarity settings in Appendix~\ref{app:benchmark-splits}: SearchQA and DocVQA use cosine similarity between BGE question embeddings, while LiveMath uses TF--IDF cosine similarity. DocVQA retains question-type constraints and LiveMath retains fine-grained mathematical-type constraints. Similarity is computed from question text only. Candidates violating these constraints receive $-\infty$ in the ranking expression above.

Let $A(q_i^{(u)};G^*)$ denote the active nodes in the final graph that were created or modified by accepted patches citing that update question. The rendered node set is:
\[
R_{\mathrm{NQ}}(q)=A(q_{i^*}^{(u)};G^*).
\]
This variant retrieves by question-to-question similarity and presents only nodes associated with the nearest update question, without adding other nodes. The selected set can be empty or contain multiple nodes. It evaluates whether skills derived from update cases transfer to test questions.

\paragraph{Retrieval records.}
For each question, we record the selected nodes, similarity scores, patch provenance, prompt length, and evaluation order. The graph, generation settings, and original multimodal input are fixed according to the protocol above.

\subsection{Update-Mechanism Ablations}
\label{app:update-ablation}

The four ablations use the same update pool, student model, self-teacher setting, and three-epoch schedule as GraphSkillAA (Full). \emph{No structured attribution} replaces structured attribution with a generic failure summary while using the same rules for constructing evidence and retaining the teacher, joint synthesis, and both Gates. \emph{No correct-case protection} removes successful examples as contrasts. \emph{No Local Gate} skips affected-case retesting and sends candidate edits directly to merging and whole-graph validation. \emph{No Big Gate} directly commits locally accepted edits.

\subsection{Teacher-Transfer Evaluation}
\label{app:teacher-ablation}

Teacher transfer holds the student, data, edit rules, three-epoch schedule, and evaluation fixed while changing only the teacher. We report three independent optimization-and-evaluation runs. GPT-5.4-mini and GPT-5.6-sol each propose edits for the two student models, yielding two self-teacher and two cross-model pairings.

\subsection{Comparison Protocol for Table~\ref{tab:table-c}}
\label{app:matched-ablation}

Colored superscripts in Table~\ref{tab:table-c} report gains over Flat skill. Component ablations use each student's self-teacher GraphSkillAA (Full) as the reference. We subtract each ablation result from its matched Full result and take the arithmetic mean of the six model--task differences. Table~\ref{tab:teacher-matched-reading} lists the references in SearchQA/LiveMath/DocVQA order. Retrieval interventions keep the same self-teacher final graph fixed and change visible content only at evaluation.

\begin{table}[ht]
\centering
\footnotesize
\renewcommand{\arraystretch}{1.16}
\renewcommand{\tabularxcolumn}[1]{m{#1}}
\begin{tabularx}{0.92\textwidth}{>{\raggedright\arraybackslash}m{0.18\textwidth}>{\raggedright\arraybackslash}m{0.13\textwidth}>{\raggedright\arraybackslash}m{0.16\textwidth}>{\raggedright\arraybackslash}X}
\toprule
\rowcolor{tableheadergray}
\multicolumn{1}{c}{\textbf{Rows interpreted}} & \multicolumn{1}{c}{\textbf{Student}} & \multicolumn{1}{c}{\textbf{Teacher}} & \multicolumn{1}{c}{\textbf{Matched reference}} \\
\midrule
Graph, retrieval, and update variants & GPT-5.6-sol & GPT-5.6-sol & GraphSkillAA (Full): 81.5 / 66.7 / 91.2. \\
\lightrowrule
Graph, retrieval, and update variants & GPT-5.4-mini & GPT-5.4-mini & GraphSkillAA (Full): 75.7 / 23.9 / 86.3. \\
\lightrowrule
Teacher-transfer evaluation & Either listed student & GPT-5.4-mini vs. GPT-5.6-sol & The two GraphSkillAA (Full) rows for that fixed student; every non-teacher setting is identical. \\
\bottomrule
\end{tabularx}
\caption{References for interpreting Table~\ref{tab:table-c}.}
\label{tab:teacher-matched-reading}
\end{table}

For example, the six differences for No structured attribution are 8.3, 28.2, 4.5, 2.7, 5.4, and 0.6, averaging 8.3 percentage points.

Changing the teacher from GPT-5.4-mini to GPT-5.6-sol improves the GPT-5.6-sol student by 3.3, 2.0, and 1.2 percentage points on SearchQA, LiveMath, and DocVQA, respectively. The same teacher change improves the GPT-5.4-mini student by 2.5, 5.2, and 2.7 percentage points.

\subsection{Shuffling Initial Edit Targets}
\label{app:shuffled-attribution}

\paragraph{Controlled setup.}
To test how initial localization affects repair, we compare Original Attribution with Shuffled Attribution. Original Attribution uses the nodes, fields, or relations selected by GraphSkillAA's attribution module. Shuffled Attribution randomly replaces those targets with other type-compatible graph objects after initial localization and before joint synthesis. For example, a target \texttt{how\_to\_use} field is replaced by the same field of another node. Both conditions use the same data splits, initial graph, student and teacher models, evidence-construction procedure, editing prompt templates, patch-generation budget, and two-level Gate rules. In each epoch, failed examples, successful contrasts, and execution evidence are derived from executions of each condition's current graph. Target shuffling changes only the edit targets. Operation types, graph-structure constraints, and subsequent validation rules remain unchanged.

GPT-5.6-sol serves as both student and teacher. The update/test split sizes are 800/200 for SearchQA, 468/117 for LiveMath, and 800/200 for DocVQA. Both conditions start from G0 + Trace/Usage and run independently with seeds 42, 43, and 44, with three optimization epochs per run.

\paragraph{Metrics and results.}
For each seed, we compare the initial G0 + Trace/Usage graph with the final committed graph example by example on the complete update pool. Fixed counts incorrect-to-correct changes, Broken counts correct-to-incorrect changes, and Net Gain is their difference. Counts are summed across the three seeds. Test Accuracy is evaluated on the respective independent test sets and reported as the mean and half-range across the three seeds.

\begin{table}[H]
\centering
\small
\renewcommand{\arraystretch}{1.14}
\begin{tabularx}{\textwidth}{@{}l>{\raggedright\arraybackslash}Xrr@{}}
\toprule
\rowcolor{tableheadergray}
\textbf{Dataset} & \textbf{Metric} & \makecell[r]{\textbf{Original}\\\textbf{Attribution}} & \makecell[r]{\textbf{Shuffled}\\\textbf{Attribution}} \\
\midrule
SearchQA & Fixed & 240 & 200 \\
 & Broken & 112 & 100 \\
 & Net Gain & \textbf{+128} & +100 \\
 & Test Accuracy (\%) & \textbf{\hrange{81.5}{1.3}} & \hrange{78.3}{1.3} \\
\lightrowrule
LiveMath & Fixed & 350 & 260 \\
 & Broken & 134 & 150 \\
 & Net Gain & \textbf{+216} & +110 \\
 & Test Accuracy (\%) & \textbf{\hrange{66.7}{2.1}} & \hrange{60.7}{2.1} \\
\lightrowrule
DocVQA & Fixed & 105 & 75 \\
 & Broken & 33 & 35 \\
 & Net Gain & \textbf{+72} & +40 \\
 & Test Accuracy (\%) & \textbf{\hrange{91.2}{1.0}} & \hrange{89.7}{1.3} \\
\bottomrule
\end{tabularx}
\caption{Original attribution versus target shuffling. Counts are summed across three seeds; test accuracy is reported as mean $\pm$ half-range.}
\label{tab:shuffled-attribution}
\end{table}

Original Attribution fixes 240, 350, and 105 cases on the three tasks, compared with 200, 260, and 75 for shuffled targets. Net gains are 128 versus 100, 216 versus 110, and 72 versus 40. On SearchQA, Original Attribution breaks 12 more cases but fixes 40 more, yielding 28 additional net corrections. On LiveMath and DocVQA, it both fixes more cases and breaks fewer. Test accuracy is higher by 3.2, 6.0, and 1.5 percentage points, respectively.

Under the same optimization settings, diagnosed targets yield greater net repair gains and higher test accuracy than random type-compatible targets. In Table~\ref{tab:table-c}, structured attribution also improves over generic failure summaries by 8.3 percentage points on average across six model--task settings, further supporting repair directed at specific skill objects.

\subsection{Retrieval Coverage of Used Nodes and Relations}
\label{app:retrieval-coverage}

To examine whether S--Q retrieval omits skills and relations used in Full execution, we compare Top-6 retrieval with GraphSkillAA (Full)'s usage records on the same optimized graph. The model is GPT-5.6-sol, with original test inputs and inference settings held fixed. Six nodes are retrieved per question, and an edge is retained only when both endpoints are selected.

For test question $q$, let $R_V(q)$ and $R_E(q)$ denote retrieved nodes and retained edges, and let $U_V(q)$ and $U_E(q)$ denote nodes and edges in Full's usage record. Sets contain distinct object IDs. Per-question coverage is defined as
\[
\operatorname{NodeCoverage}(q)=\frac{|R_V(q)\cap U_V(q)|}{|U_V(q)|},\qquad
\operatorname{EdgeCoverage}(q)=\frac{|R_E(q)\cap U_E(q)|}{|U_E(q)|}.
\]
Node and edge coverage are macro-averaged separately over questions with nonzero denominators. Define
\[
I_V=\{q:|U_V(q)|>0\},\qquad I_E=\{q:|U_E(q)|>0\},
\qquad N_V=|I_V|,\quad N_E=|I_E|.
\]
The mean coverages reported in Table~\ref{tab:retrieval-coverage} are
\[
\overline C_V=\frac{1}{N_V}\sum_{q\in I_V}\operatorname{NodeCoverage}(q),\qquad
\overline C_E=\frac{1}{N_E}\sum_{q\in I_E}\operatorname{EdgeCoverage}(q).
\]
A question with logged nodes but no logged edges contributes to node coverage only. Zero-denominator questions are excluded only from the corresponding coverage mean and remain in accuracy and other metrics. Each coverage ratio is computed per question before averaging, rather than pooling nodes or edges across questions.

\begin{table}[H]
\centering
\small
\renewcommand{\arraystretch}{1.14}
\setlength{\tabcolsep}{3pt}
\begin{tabularx}{\textwidth}{@{}l*{5}{>{\raggedleft\arraybackslash}X}@{}}
\toprule
\rowcolor{tableheadergray}
\textbf{Dataset}
& \makecell[r]{\textbf{Retrieved}\\\textbf{Nodes}}
& \makecell[r]{\textbf{Node}\\\textbf{Coverage}}
& \makecell[r]{\textbf{Edge}\\\textbf{Coverage}}
& \makecell[r]{\textbf{S--Q}\\\textbf{Accuracy}}
& \makecell[r]{\textbf{GraphSkillAA}\\\textbf{(Full)}\\\textbf{Accuracy}} \\
\midrule
SearchQA & 6 & 84.6\% & 72.3\% & 73.5\% & 81.5\% \\
\lightrowrule
LiveMath & 6 & 69.8\% & 56.4\% & 42.5\% & 66.7\% \\
\lightrowrule
DocVQA & 6 & 88.2\% & 78.6\% & 86.0\% & 91.2\% \\
\bottomrule
\end{tabularx}
\caption{Mean per-question coverage of nodes and edges in Full's usage records by S--Q retrieval. Accuracy repeats the GPT-5.6-sol results from Table~\ref{tab:table-c}.}
\label{tab:retrieval-coverage}
\end{table}

Top-6 retrieval omits, on average, 15.4\%, 30.2\%, and 11.8\% of Full-logged nodes and 27.7\%, 43.6\%, and 21.4\% of Full-logged edges. Edge coverage is lower than node coverage on all three tasks. Only edges whose endpoints are both selected survive retrieval, so edges incident to omitted nodes are also removed. LiveMath has the lowest node and edge coverage and the largest accuracy gap, 24.2 percentage points. The gaps for SearchQA and DocVQA are 8.0 and 5.2 percentage points. These results show that selecting a fixed number of skills by question similarity omits some rules and coordinating relations used during full-graph execution.

To distinguish information coverage from relations and activation conditions, Table~\ref{tab:table-c} provides two controls that retain all nodes. Nodes only keeps every node and field while removing edges, reducing accuracy by 4.0, 4.6, and 1.2 percentage points. Hiding only \texttt{when\_to\_use} and \texttt{avoid}, with all nodes and edges retained, reduces accuracy by 11.0, 24.0, and 6.9 percentage points. Coverage measures which skill objects retrieval retains, Nodes only examines relations, and field hiding examines applicability and exclusion conditions. The Full-versus-S--Q comparison evaluates complete-graph presentation against Top-6 retrieval.

\section{Residual Error Analysis}
\label{app:badcases}

We select two failed test cases from each benchmark and compare the question, model answer, reference answer, and execution record to examine what went wrong and how it could be addressed.

\begin{badcasebox}{SearchQA: scoring mismatch versus missing ordinal evidence}
\textbf{Correct person, rejected name form.} For the clue about the leader convicted by Panamanian courts in 1993, the model answers \emph{Manuel Noriega}, while the reference is \emph{Noriega}. Both identify the same person, but strict matching marks the full name as incorrect. Alias normalization or a separately reported semantic score could address this issue.\par
\medskip
\textbf{Evidence of holding office, but not of being first.} Asked for the first U.S. Secretary of State, the model answers \emph{James Madison}, while the reference is \emph{Thomas Jefferson}. The retrieved text states only that Madison served as Secretary of State while Jefferson was president. It does not identify the first holder of the office. The model selects Madison from this appointment information. Improvement requires evidence for ``first'' and a check of whether the available material is sufficient to answer the question.
\end{badcasebox}

\begin{badcasebox}{LiveMath: theorem identification versus realizability}
\textbf{Related conclusion mistaken for the requested equivalence.} The model chooses a co-$t$-structure characterization (E), while the reference is restriction of the induced $t$-structure to $T_c^-$ (B). Although the model checks the hypotheses and options, it treats a related consequence as an equivalent condition. The theorem's exact statement and assumptions need to be checked.\par
\medskip
\textbf{Necessary conditions satisfied, but realizability unproved.} The model selects C, which includes polygon type $\{6,10\}$, while reference A permits only $\{4,12\}$ and $\{8,8\}$. The model correctly derives that both side counts are even, sum to 16, and cannot equal 2, but then assumes that every remaining even partition is realizable. The missing step is a construction or an obstruction check, which could be supplied by consulting a classification theorem or explicitly checking realizability.
\end{badcasebox}

\begin{badcasebox}{DocVQA: character recognition versus role selection}
\textbf{Correct region, wrong handwritten date.} The model finds the date in the memo but reads \texttt{11/18/82} as \texttt{7/18/82}. Later extraction and formatting steps retain this reading. Improvement should start with recognizing the date again, for example by enlarging the image region or cross-checking with OCR.\par
\medskip
\textbf{A role requested, but a name returned.} For the question \emph{Who need to sign at bottom of page?}, the model returns the handwritten name \emph{Lillian B Dalton}, while the reference expects \emph{Chief, Employment Branch}. Both the name and role appear in the execution record, but the model interprets the question as asking who signed and fails to distinguish the signer from the office required to sign. If this error recurs, the skill could specify when to select a name and when to select a role.
\end{badcasebox}

Different errors call for different remedies: name variants for scorer normalization, missing facts or theorems for retrieval, misread characters for image rereading or OCR, and recurring rule-selection errors for skill edits. GraphSkillAA returns \texttt{NO\_PATCH} when it finds no reusable skill defect. A future extension could route such errors to the appropriate tools.

\section{Discussion, Limitations, and Future Work}
\label{app:discussion}

\subsection{Interpretation of Experimental Findings}
\label{app:interpretation}

Node fields help the model decide when to use a rule, while edges describe how rules work together. Optimized nodes retain substantial gains after edges are removed, showing the value of node content itself. Hiding applicability and exclusion conditions causes a larger average loss, highlighting the importance of using rules in the appropriate context.

Node and edge IDs also connect the update stages: attribution specifies edit locations, repair groups store the changes, and the Gates use affected examples to decide which changes to retain or revert. Ablations show gains from both structured attribution and local retesting. The cases in Appendix~\ref{app:badcases} show how skill repair can be combined with improvements in knowledge retrieval, visual recognition, and scoring.

The ALFWorld results suggest using performance under a longer step budget to investigate why a task did not finish within the original budget. GraphSkillAA edits the graph when it finds a skill defect and otherwise retains existing rules through \texttt{NO\_PATCH}.

\subsection{Limitations}
\label{sec:limitations}

\paragraph{Skill-library size and task length.}
The current skill graph fits in the context in full. Larger libraries increase prompt cost, while errors at the end of longer tasks are harder to trace to specific steps. Extending to these settings requires suitable skill-selection and attribution mechanisms.

\paragraph{Edit granularity.}
The current system uses node fields and edges as edit addresses, although an edit may change only one sentence or fragment within a field. Future versions could assign separate identifiers to sentences, theorems, table cells, or image regions for finer localization, while retaining paired retesting and rollback at the repair-group level.

\paragraph{Repair strategies and evaluation feedback.}
The current update target is the skill graph. Future systems could select retrieval, perception tools, model training, or skill edits according to the error type, and improve scoring feedback through alias recognition and semantic-equivalence judgments. These choices also need evaluation under distribution shifts.

\subsection{Long-Term Research Directions}
\label{app:research-directions}

\paragraph{Skill selection at scale.}
Selecting Top-$K$ rules by similarity alone can omit prerequisites or retrieve an inapplicable procedure. Graph-of-Skills, SkillDAG, HyperSkill, and CaSKG already use dependencies and higher-order structure~\citep{graphofskills2026,skilldag2026,hyperskill2026,caskg2026}. These ideas could support joint optimization of rule selection, dependency completion, and context cost, with checks for missing steps and conflicts in the selected subgraph.

\paragraph{Testing attribution through intervention.}
Temporarily removing, replacing, or adjusting nodes and edges and observing the outcome can test individual contributions and interactions~\citep{caskg2026,demystifyingskills2026}. In longer tasks, these comparisons may also help trace a final error to earlier steps.

\paragraph{Ongoing skill-library maintenance.}
Skill libraries need to merge, split, specialize, retire, or restore rules according to use. SkillOS, SkillsBench, and boundary-aware memory offer ideas for version tracking, conflict tests, pruning low-value skills, and reviewing imported skills~\citep{skillos2026,skillsbench2026,boundaryskill2026,agentskillssurvey2026}.

\paragraph{Joint learning of models and external skills.}
External skills are easy to inspect and revise, while model training may make familiar procedures more reliable. Frequently changing procedures could remain in the graph, repeatedly validated skills could provide training material, and the updated model could then propose and test further skills~\citep{trace2skill2026,reasoningbank2025,memskill2026,evoskills2026}.

\end{document}